\documentclass[preprint,12pt,authoryear]{elsarticle}

\usepackage[utf8]{inputenc}
\usepackage[T1]{fontenc}
\usepackage{amsmath}
\usepackage{amssymb}
\usepackage{tikz}
\usetikzlibrary{positioning,arrows.meta,calc,fit}
\usepackage{pgfplots}
\pgfplotsset{compat=1.18}
\usepgfplotslibrary{groupplots}
\usepackage{multirow}
\usepackage{diagbox}
\usepackage{booktabs}
\usepackage{microtype}
\usepackage{xurl}
\usepackage[section]{placeins}
\usepackage{algorithm}
\usepackage{algpseudocode}
\AddToHook{env/table/begin}{%
  \setlength{\abovecaptionskip}{0pt}%
  \setlength{\belowcaptionskip}{4pt}%
}
\AddToHook{env/table*/begin}{%
  \setlength{\abovecaptionskip}{0pt}%
  \setlength{\belowcaptionskip}{4pt}%
}

\journal{Computers \& Operations Research}

\begin{document}

\begin{frontmatter}

\title{GNN-Guided Network Reduction for Centralized Urban Traffic Routing}

\author[inst1,inst2]{Nikolai Antonov\corref{cor1}}
\ead{antonni1@fel.cvut.cz}

\author[inst1]{Lukáš Chrpa}

\author[inst1,inst2]{Andrii Nyporko}

\author[inst1]{Jan Hůla}

\cortext[cor1]{Corresponding author}

\affiliation[inst1]{organization={Czech Technical University in Prague, Czech Institute of Informatics, Robotics and Cybernetics},
  addressline={Jugoslávských partyzánů 1580/3},
  city={Prague},
  postcode={160 00},
  country={Czech Republic}}

\affiliation[inst2]{organization={Czech Technical University in Prague, Faculty of Electrical Engineering},
  addressline={Technická 2},
  city={Prague},
  postcode={166 27},
  country={Czech Republic}}

\begin{abstract}
We consider the centralized traffic routing (CTR) problem, where a central planner assigns routes to origin-destination (OD) vehicle flows to minimize congestion-dependent segment cost. Prior work formulates CTR efficiently as a mixed-integer linear program (MILP), but scaling this formulation to the full road network is computationally prohibitive within an operational time budget. Bounded Routes and Smart Routes speed up optimization using precomputed reduced networks; however, these methods do not account for vehicle information within each OD flow. Motivated by this gap, we propose a learning-based urban network reduction approach. A dedicated graph neural network (GNN) first scores road segments for each OD flow using the road network and vehicle-flow information. Highly ranked segments are then incorporated into complete OD routes, and the original segment-flow MILP is solved on the resulting reduced road network. The GNN thus guides the reduction of the road network, while the solver enforces feasibility and determines the routing. Experiments on four real urban networks -- Prague-Dejvice, London, New York, and Sydney -- show that the proposed approach achieves the smallest relative objective difference to the $30$-second full-graph reference in every city under both traffic regimes. Averaged across cities, this difference is more than six times lower than that of the best competing method. Moreover, traffic simulation results show that the optimization gains of the proposed method lead to more efficient routing, with the highest aggregate arrival rate and a competitive trip duration per arrival percentage point.
\end{abstract}


\end{frontmatter}


\section{Introduction}
\label{sec:introduction}

Traffic congestion in urban road networks imposes substantial economic \citep{Serok2022,GaanCardenas2024,Gunjo2024} and environmental~\citep{Khreis2024} costs, making urban traffic management a core objective for intelligent transportation systems. Unlike decentralized navigation, in which each driver selects a route independently, \emph{centralized traffic routing} (CTR) jointly assigns routes to all vehicles simultaneously. Such coordination can reduce travel time, as studied both theoretically~\citep{Cominetti2021} and using real mobility data~\citep{Monnot2022}. In an operational setting, drivers submit their destinations and receive coordinated routes from a central planner. At each planning step (episode), the planner processes the current set of requests and must compute routes jointly for all vehicles on a large road network within a short operational time budget, which poses a challenging computational task.

Our prior work introduced a segment-flow mixed-integer linear programming (MILP) formulation for CTR and showed its advantages over planning-based approaches~\citep{Nyporko2025,Nyporko2026}. Despite the demonstrated efficiency, solving this MILP within a short time budget becomes increasingly difficult as the road network and the number of vehicle flows grow. A common way to accelerate solving in this case is to reduce the road network before optimization. Prominent fixed-reduction methods -- Bounded Routes (BR) and Smart Routes (SR)~\citep{svadlenka2023_1} -- precompute a pool of routes for each origin-destination (OD) pair and use the union of their segments as a reduced network across episodes. However, this creates a trade-off: conservative reduction may leave the MILP size close to the full-graph formulation, whereas aggressive pruning can remove valuable alternatives needed for low-cost routing.

Since historical traffic data reveal regularities in road congestion~\citep{Wang2015}, we come up with the idea of tackling CTR using a data-driven approach. In particular, we propose an urban network reduction guided by a machine learning model that evaluates the road segments based not only on the network structure, but also on the information about OD vehicle flows. For each episode, a dedicated graph neural network (GNN) scores segments separately for every OD flow and ranks them by predicted relevance for routing. Rather than predicting final routes or exact segment flows, these rankings guide the construction of an episode-specific reduced network, which is then used to restrict the original segment-flow MILP. Thus, the GNN accelerates solving without taking over the optimization model's role: it identifies promising parts of the road network, while the solver enforces model feasibility and determines the final routing.

We evaluate the proposed approach on real road network topologies from Prague-Dejvice, London, New York, and Sydney, using scenarios with different numbers of OD flows and two traffic regimes. On these scenarios, we first compare our GNN-guided reduction with fixed-reduction baselines in terms of solution quality and runtime under the short operational solver budget of $10$ seconds. To assess the contribution of learned ranking, we also test the same reduction procedure with non-learned ranking alternatives. In these comparisons, solution quality is measured relative to a $30$-second run of the CTR MILP on the full road network, using the full episode length. Finally, we examine whether the differences observed at the optimization level carry over to traffic by replaying the resulting routes in the simulator of urban mobility (SUMO)~\citep{SUMO2018} and measuring arrival rate and trip duration relative to the achieved arrival rate. 

Our contributions are as follows:
\begin{itemize}

\item \textbf{A ranking-guided network reduction for CTR.}
We propose an algorithm that uses per-flow segment rankings to construct a reduced road network for each CTR episode. By connecting selected segments through complete OD paths and locally expanding their union, the algorithm reduces the number of routing variables in the original segment-flow MILP.

\item \textbf{A dedicated GNN for learning flow-specific segment rankings.}
To guide the road network reduction, we introduce a GNN that combines the full MILP representation, road-network characteristics, and current OD flows to rank segments separately for each flow. The model learns segment orderings from labelled routing solutions rather than predicting exact vehicle counts or final routes, while the solver determines the final routing and enforces feasibility. For each city and traffic regime, a single trained model generalizes to unseen scenarios with different numbers of OD flows.

\item \textbf{The best solution quality under a short solver budget.}
We evaluate the resulting GNN-guided reduction on four real urban road networks, comparing it with fixed-reduction baselines and with the same procedure using non-learned rankings. The proposed GNN-guided network reduction approach achieves the best solution quality in terms of the smallest mean relative objective difference compared to other methods in every city and traffic regime.

\item \textbf{Route recovery and simulation-level validation.}
To transform MILP solutions into individual vehicle routes, we present a dedicated route-recovery procedure and then evaluate the recovered routes in SUMO under different shares of rerouted vehicles to assess whether the optimization gains lead to improved traffic outcomes. Under the considered stress-test scenarios, our proposed approach yields the highest overall arrival rate across cities in both traffic regimes and remains competitive in trip duration relative to the achieved arrival rate.

\end{itemize}

The remainder of the paper is organized as follows. Section~\ref{sec:related_work} reviews centralized traffic routing, network reduction, and learning-assisted optimization. Section~\ref{sec:problem} formulates the considered CTR problem as a MILP with turn restrictions. Section~\ref{sec:methodology} describes the fixed reduction baselines and our proposed ranking-guided reduction framework, including GNN-based segment ranking and route recovery. Section~\ref{sec:experiments} presents the experimental setup and evaluates scalability, ranking quality, and performance in optimization and microscopic simulation. Section~\ref{sec:conclusion} concludes the work.

\section{Related Work}
\label{sec:related_work}

In this section, we first review decentralized, partially centralized, and centralized routing to place CTR within the broader literature (Section~\ref{subsec:related_decent_vs_cent}). We then review static and dynamic CTR alongside planning-based approaches (Section~\ref{subsec:related_taxonomy}), emphasizing the role of mixed-integer CTR formulations. Finally, we discuss the network reduction methods (Section~\ref{subsec:related_preprocess}) and connect them with learning-assisted routing and optimization approaches most closely related to the proposed GNN-guided reduction (Section~\ref{subsec:related_learning}).

\subsection{Decentralized, partially centralized, and centralized routing}
\label{subsec:related_decent_vs_cent}

When drivers choose routes independently, the resulting traffic pattern corresponds to a user equilibrium, which is generally less efficient than the system-optimal routing selected by a central planner~\citep{Roughgarden2002}. The relative difference between decentralized and system-optimal routing tends to be most pronounced at intermediate traffic levels~\citep{ColiniBaldeschi2020}, while empirical studies on real road networks show that uncoordinated drivers can experience considerably longer travel times~\citep{Youn2008,Zhang2018}. Thus, decentralized routing cannot generally be expected to achieve system-optimal performance.

Several partially centralized approaches aim to adjust the behavior of decentralized systems. Information-based approaches improve drivers' visibility of road network conditions through real-time guidance and navigation apps, so that selfish routing avoids overused roads~\citep{Bagabaldo2024}. Incentive-based approaches modify route choice through congestion pricing~\citep{Maheshwari2024} or controlled information release~\citep{Ambrogio2024}, aligning selfish best-response with the system optimum. Compliance-aware approaches design system-optimal recommendations that remain effective when only a fraction of drivers follow them~\citep{Yun2024}. Thus, partial centralization can narrow the gap between decentralized and centralized routing, but its effectiveness depends on how drivers respond to information, incentives, and route recommendations.

Unlike partially centralized approaches, a centralized planner jointly assigns routes to all vehicles to minimize the total travel cost across the road network~\citep{Morandi2023}. This requires accounting for their interactions and the resulting congestion. However, the optimization problem becomes increasingly difficult as the road network and traffic volumes grow~\citep{Chrpa2023}, while practical applications require high-quality solutions within limited computation time. Thus, centralized routing directly targets system-optimal performance, but at a substantial computational cost. The following subsection reviews the main CTR formulations and solution approaches developed for this setting.

\subsection{CTR formulations and static mixed-integer optimization}
\label{subsec:related_taxonomy}

CTR is commonly studied within the framework of static traffic assignment, where routes are computed for a given number of vehicles traveling between specified origins and destinations. A representative formulation is the segment-flow MILP of \citet{Nyporko2025}, which assigns integer vehicle flows to road segments and minimizes a congestion-dependent objective. Related formulations adapt the system-optimal model to practical requirements. For example, \citet{Jahn2005} introduce user constraints that limit how much longer an assigned route may be than the shortest route available to that user, while \citet{Angelelli2021} use a piecewise-linear approximation to obtain a linear formulation. A broader overview of static traffic assignment formulations is provided by \citet{Morandi2023}. Thus, static CTR formulations can combine system-wide optimization with practical user-level requirements. However, the works reviewed above primarily contribute by introducing new CTR formulations or extending existing ones, while our work addresses a complementary computational question: how to reduce the road network to solve a given segment-flow MILP more efficiently.

Dynamic traffic assignment (DTA) extends the static setting by modeling how traffic flows and travel times evolve over time. Existing system-optimal approaches include link-based formulations~\citep{Long2019}, path-based formulations~\citep{Zhang2020}, and models combining routing with traffic flow control~\citep{Como2016}. Unlike these approaches, our formulation solves each episode as a static routing problem, while microscopic simulation is used only to evaluate the resulting routes; therefore, DTA lies outside the scope of this paper.

A separate line of work addresses centralized routing through automated planning. \citet{ChrpaVP19} formulate CTR as a planning problem in which the generated plan defines a congestion-mitigating routing for all vehicles. This framework was evaluated in microscopic traffic simulation~\citep{Vallati2020} and later extended to hybrid and real-time settings~\citep{Vallati2021}. Our previous research~\citep{Nyporko2025} compared the planning-based methods with a segment-flow MILP and found that the MILP outperformed them in scalability and microscopic simulation performance. We therefore continue to study CTR through the mixed-integer segment-flow formulation.

\subsection{Road network reduction for mixed-integer CTR}
\label{subsec:related_preprocess}

The main obstacle to solving CTR through mixed-integer programming is the limited scalability of the resulting MILP model. Existing approaches address this issue either by accelerating the solver~\citep{Zhang2023} or by reducing the instance size. The latter strategy is commonly implemented through route-based preprocessing that constructs a reduced road network before optimization~\citep{svadlenka2023_1}. Prominent methods in this line of work include Bounded Routes (BR) and Smart Routes (SR)~\citep{svadlenka2023_1}, forming a part of a broader framework for intelligent urban traffic routing~\citep{svadlenka2023_2}. BR enumerates routes whose lengths are bounded relative to the shortest route, while SR additionally clusters their segment sets using the Jaccard distance and retains only representative alternatives. Our work also follows this general direction and focuses on reducing the road network before solving the MILP rather than modifying the solver itself.

A key limitation of BR and SR is that the reduced road network remains fixed across episodes~\citep{svadlenka2023_1}. However, the contribution of individual segments to routing quality depends on the OD flows in the current episode: a segment that is unnecessary in one episode may provide an important routing alternative in another. Road network vulnerability studies show that the impact of segment disruptions on total travel cost depends on traffic volumes and congestion, while network structure alone may not identify the most critical segments~\citep{Chen2012,Gauthier2018}. Our previous experiments similarly found that the stronger pruning of SR produced longer simulated trip durations than BR, indicating that removing too many alternatives can introduce a significant suboptimality~\citep{Nyporko2025}. Thus, fixed preprocessing creates a trade-off between reducing the MILP and preserving routes that may be useful in the current episode.

\subsection{Learning-assisted optimization for CTR}
\label{subsec:related_learning}

Combinatorial optimization is known to incorporate machine learning as heuristics for computationally hard problems~\citep{Bengio2021,Angioni2025}. In traffic routing, learned models can use information about the OD flows in the current episode to provide better routing decisions. For example, reinforcement learning approaches produce routing policies for a given set of OD flows: their training can be guided by the method of successive averages~\citep{Wang2025}, or by teacher policies derived from physics-based traffic flow models~\citep{Ke2025}. Other methods use surrogate models to approximate traffic flows~\citep{Lassen2025} or estimate congestion caused by road renovations~\citep{Bosch2025}. However, the routes predicted by a learned model may violate the constraints of the original problem unless feasibility is enforced separately. A useful alternative is to use machine learning for guidance, for example, to identify informative variable subsets that simplify mixed-integer optimization~\citep{Dilkina2009,Ferber2022} or to select heuristics for modifying vehicle routes during a search~\citep{Johnn2024}. These directions suggest using learning to guide the optimization rather than to predict the final routing directly.

Following this idea, machine learning has been applied to reduce the problem size in crew rostering~\citep{Quesnel2022} and electric bus scheduling~\citep{Gerbaux2025}. \citet{Quesnel2022} use a neural network to rank pairings for each crew member and retain the most promising ones for generating schedules within a branch-and-price algorithm. In a transportation setting, \citet{Gerbaux2025} combine GNN-based and greedy arc selection to construct reduced problems within a column-generation heuristic. In both cases, learning identifies a smaller set of promising elements, while the downstream optimization procedure constructs the final solution. However, both approaches are tailored to their specific scheduling problems and the corresponding column-generation or branch-and-price frameworks. Our approach follows the same general principle but applies learned reduction directly to the variables of the segment-flow CTR MILP. It combines neural scoring of graph elements~\citep{joshi2019tsp_gcn} with learning-assisted mixed-integer reduction~\citep{liu2024lpmip}. Before optimization, the GNN processes the complete MILP instance, including its variables, constraints, objective, and current OD vehicle flows, and scores every flow-segment variable $x_r^i$. The scores though do not constitute a solution; they just estimate which routing variables should be retained. The subsequent ranking-guided procedure converts them into flow-specific segment sets, after which the original MILP is constructed and solved on the reduced network.

\section{Problem formulation}
\label{sec:problem}

We consider a centralized traffic routing problem on a congested urban road network. The goal is to jointly assign routes to all vehicles while minimizing congestion-dependent travel cost. The formulation builds on our previous segment-flow MILP~\citep{Nyporko2025}. It retains the aggregate OD-flow representation and tiered segment costs, while adding constraints that enforce the turn restrictions.

\subsection{Network, OD flows, and congestion model}
\label{subsec:network_and_congestion}

The road network is modeled as a directed graph $G=(V,R)$, where $V$ is the set of junctions and $R$ is the set of directed road segments. A single segment aggregates the lanes and parallel roadways connecting the same two junctions in the same direction, so at most one segment connects any ordered pair of nodes. We therefore identify each $r\in R$ with its endpoint pair and write $r=(u,v)$ for the segment from $u$ to $v$. Each segment has a length $\ell_r$ and an estimated travel time $\tau_r$. For $v\in V$, let $\delta^+(v)=\{(v,w)\in R\}$ and $\delta^-(v)=\{(u,v)\in R\}$ denote the outgoing and incoming segments at $v$.

Some junctions are assumed to have turn restrictions. Let $V_T\subseteq V$ denote the set of such junctions. For each $v\in V_T$, let $T_v\subseteq\delta^-(v)\times\delta^+(v)$ denote the allowed turns through $v$. A pair $(a,b)\in T_v$ means that vehicles on segment $a$ may continue through $v$ onto segment $b$. We write $T=\{T_v\}_{v\in V_T}$ for the collection of turn restrictions. Junctions outside $V_T$ allow all turn pairs.

A CTR instance is specified by a set of OD vehicle flows, which we call an \emph{episode} or, equivalently, a problem \emph{instance}. Within an episode, each flow $i\in\{1,\dots,N_F\}$ represents a group of identical vehicles traveling from origin $o^i\in V$ to destination $d^i\in V$, where $n^i\in\mathbb{N}$ denotes the number of vehicles in the group. The set $R_i\subseteq R$ contains the road segments available for routing flow $i$. Similar to~\citet{Nyporko2025}, we adopt a static, episode-level CTR setting in which every episode is solved independently of the others, the segment costs are not updated as traffic conditions change, and traffic is modeled at the level of aggregate OD flows rather than individual vehicle trajectories.

Each road segment $r\in R$ is associated with a load $s_r$, defined as the number of vehicles routed through this segment. There is no hard restriction on the segment capacity (any number of vehicles may traverse it); instead, the segment cost increases with its load according to a piecewise-linear function with upward jumps at the tier thresholds. Let $\mathcal H=\{1,\ldots,H\}$ denote the set of congestion tiers. For the first $H-1$ tiers, $\theta_r^k$ denotes the upper load threshold, with $0<\theta_r^1<\cdots<\theta_r^{H-1}$, while the final tier extends to $+\infty$. Each tier has a unit cost $c_r^k$, with $c_r^1<c_r^2<\cdots<c_r^H$. The active tier is determined by the segment load:
\begin{equation}
k(s_r)=
\begin{cases}
1, & 0\le s_r\le \theta_r^1,\\
k, & \theta_r^{k-1}<s_r\le \theta_r^k,\quad k=2,\ldots,H-1,\\
H, & s_r>\theta_r^{H-1}.
\end{cases}
\end{equation}
For a given load $s_r$, the segment cost is defined as $g_r(s_r)=c_r^{k(s_r)}s_r$, and the total cost of an episode is given by $\sum\limits_{r\in R}g_r(s_r)$.

\subsection{MILP formulation}
\label{subsec:milp_formulation}

\paragraph{Decision variables}
Table~\ref{tab:decision_variables} summarizes the decision variables.
\begin{table}[!t]
\centering
\caption{Decision variables of the CTR MILP.}
\label{tab:decision_variables}
\small
\setlength{\tabcolsep}{4pt}
\begin{tabular}{@{}l l p{0.60\linewidth}@{}}
\toprule
Variable & Domain & Description \\
\midrule
$x_r^i$ & $\{0,1,\dots,n^i\}$ & Number of vehicles of flow $i$ assigned to segment $r$ \\
$f_{a,b}^i$ & $\mathbb{R}_{\ge 0}$ & Number of vehicles of flow $i$ continuing from segment $a$ to segment $b$ through an allowed turn \\
$s_r$ & $\mathbb{R}_{\ge 0}$ & Load on segment $r$ \\
$y_r$ & $\mathbb{R}_{\ge 0}$ & Cost contribution of segment $r$ \\
$z_r^k$ & $\{0,1\}$ & Indicator that tier $k$ is active on segment $r$ \\
\bottomrule
\end{tabular}
\end{table}

We introduce three groups of variables: routing variables $x_r^i$ for segments $r\in R_i$, segment variables $s_r$, $y_r$, $z_r^k$ for segments $r\in R$, and turn-flow variables $f_{a,b}^i$ for allowed turns at junctions $v\in V_T$ other than the origin and destination of flow $i$.

\paragraph{Constraints}
Let $\bar{s}=\sum\limits_i n^i$ denote the total number of vehicles in the episode. Since the load on any segment cannot exceed $\bar{s}$, we use $M_s=\bar{s}+1$ for the upper bound of the final congestion tier and $M_y=\bar{s}\max_{r\in R,\,k\in\mathcal H}c_r^k+1$ for the cost linearization.

\begin{itemize}

\item Flow is conserved at each node:
\begin{equation}
\sum_{r\in\delta^+(v)\cap R_i} x_r^i
-\sum_{r\in\delta^-(v)\cap R_i} x_r^i
=
\begin{cases}
n^i, & v=o^i,\\
-n^i, & v=d^i,\\
0, & \text{otherwise},
\end{cases}
\quad \forall v\in V,\ \forall i.
\end{equation}

\item No flow enters an origin and none leaves a destination:
\begin{equation}
\sum_{r\in\delta^-(o^i)\cap R_i} x_r^i=0,\qquad
\sum_{r\in\delta^+(d^i)\cap R_i} x_r^i=0,
\quad \forall i.
\end{equation}

\item The load on each segment is the total number of vehicles routed through it:
\begin{equation}
s_r=\sum_{i:\,r\in R_i}x_r^i,
\quad \forall r\in R.
\end{equation}

\item Exactly one congestion tier is selected for each segment:
\begin{equation}
\sum_{k\in\mathcal H}z_r^k=1,
\quad \forall r\in R.
\label{eq:milp_one_tier}
\end{equation}

\item The segment load must satisfy the bounds of the selected tier:
\begin{equation}
\begin{aligned}
s_r
&\le
\sum_{k=1}^{H-1}\theta_r^k z_r^k+M_s z_r^H,
&& \forall r\in R,\\
s_r
&\ge
\sum_{k=2}^{H}\theta_r^{k-1}z_r^k,
&& \forall r\in R.
\end{aligned}
\label{eq:milp_tier_bounds}
\end{equation}
At an exact threshold, either adjacent tier is feasible; since unit costs increase and $y_r$ is minimized, the lower-cost tier is selected.

\item The segment cost is linearized using the selected tier:
\begin{equation}
y_r\ge c_r^k s_r-M_y(1-z_r^k),
\quad
\forall r\in R,\ \forall k\in\mathcal H.
\label{eq:milp_cost_lin}
\end{equation}

\item At junctions with turn restrictions, the per-turn flows reproduce the incoming and outgoing segment flows:
\begin{align}
\sum_{b\in R_i:\,(a,b)\in T_v} f_{a,b}^i
&=x_a^i,
&& \forall v\in V_T,
\forall a\in\delta^-(v)\cap R_i,
\forall i:\ o^i\neq v,\ d^i\neq v,\\
\sum_{a\in R_i:\,(a,b)\in T_v} f_{a,b}^i
&=x_b^i,
&& \forall v\in V_T,
\forall b\in\delta^+(v)\cap R_i,
\forall i:\ o^i\neq v,\ d^i\neq v.
\end{align}

We remark that imposing integrality on $f_{a,b}^i$ is unnecessary at this stage, as it could make the formulation harder to solve. For fixed integer segment flows $x_r^i$ from a feasible MILP solution, the turn-flow constraints at each restricted junction can be represented as a maximum-flow problem on a bipartite network whose capacities are given by integers $x_r^i$. Hence, an integer assignment to the variables $f_{a,b}^i$ exists and can be recovered in polynomial time, as described in Section~\ref{subsec:route_recovery}.

\end{itemize}

\paragraph{Objective}
The objective minimizes the total segment cost:
\begin{equation}
\min\quad
\sum_{r\in R} y_r.
\label{eq:milp_objective}
\end{equation}

\section{Solution Methodology}
\label{sec:methodology}

\begin{figure}[!t]
\centering
\begin{tikzpicture}[
state/.style={
draw, rounded corners=2pt, align=center, inner sep=3pt,
minimum height=9mm, text width=0.13\linewidth, font=\scriptsize
},
fixed/.style={state, draw=blue!65!black, fill=blue!7},
proposed/.style={state, draw=orange!80!black, fill=orange!10},
result/.style={state, draw=green!50!black, fill=green!9},
process/.style={
align=center, fill=white, inner sep=1.5pt,
font=\scriptsize\itshape, text=gray!75!black
},
ptitle/.style={font=\small\bfseries, anchor=south west},
arrow/.style={-Latex, line width=0.8pt, draw=gray!75!black},
node distance=13mm
]
\node[fixed] (f0) {network and\\OD pairs};
\node[fixed, right=24mm of f0] (f1) {segment set\\$R_i'$};
\node[result, right=24mm of f1] (f2) {routing\\solution};

\draw[arrow] (f0) -- node[process, above] {build route union} (f1);
\draw[arrow] (f1) -- node[process, above] {solve MILP} (f2);

\node[draw=blue!30, rounded corners=3pt, fit=(f0)(f1)(f2),
inner xsep=3mm, inner ysep=5mm] (fg) {};
\node[ptitle, text=blue!65!black] at ([xshift=1mm,yshift=1mm]fg.north west)
{(a) Fixed};

\node[proposed, below=20mm of f0, xshift=-12mm] (p0) {current\\episode};
\node[proposed, right=12mm of p0] (p1) {per-flow segment\\scores $q_r^i$};
\node[proposed, right=23mm of p1] (p2) {segment set\\$R_i'$};
\node[result, right=16mm of p2] (p3) {routing\\solution};

\draw[arrow] (p0) -- node[process, above] {predict} (p1);
\draw[arrow] (p1) -- node[process, above] {build path union\\and expand} (p2);
\draw[arrow] (p2) -- node[process, above] {solve MILP} (p3);

\node[draw=orange!35, rounded corners=3pt, fit=(p0)(p1)(p2)(p3),
inner xsep=3mm, inner ysep=5mm] (pg) {};
\node[ptitle, text=orange!80!black] at ([xshift=1mm,yshift=1mm]pg.north west)
{(b) Episode-specific};
\end{tikzpicture}

\caption{Fixed and episode-specific network reduction. Boxes show states, while arrow labels show operations. \textbf{(a)}~The fixed procedure constructs the flow-specific segment sets $R_i'$ independently of the current episode. \textbf{(b)}~Our approach predicts per-flow segment scores and constructs these sets for the current episode. Both approaches use the same segment-flow MILP.}
\label{fig:schemes}
\end{figure}
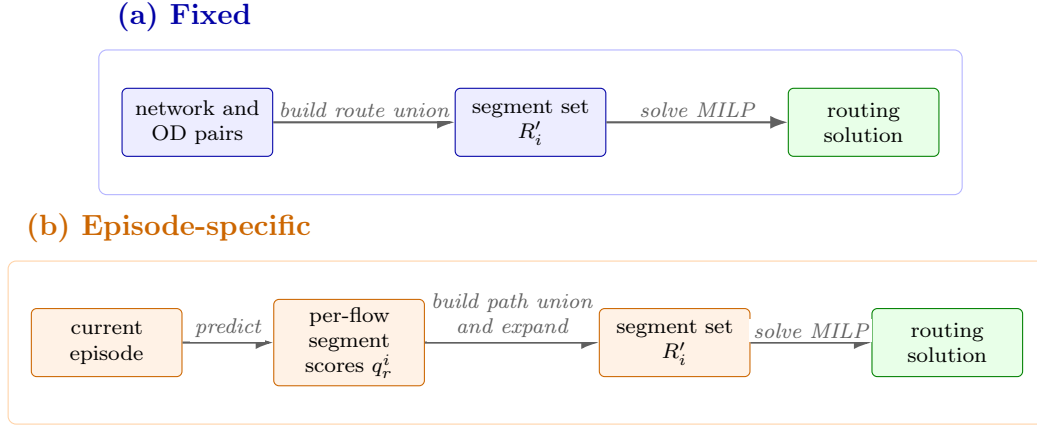

This section presents our GNN-guided network reduction approach and the fixed-reduction methods used for comparison. In every case, the goal is to restrict the segment-flow MILP by replacing the set $R_i$ available to each flow $i$ with a flow-specific segment set $R_i'\subseteq R_i$; the formulation in Section~\ref{subsec:milp_formulation} then includes routing variables and associated constraints only for segments in $R_i'$. We use the term \emph{reduced network} to refer to the resulting collection $\{R_i'\}_{i=1}^{N_F}$. However, the presented approaches differ in how the flow-specific segment sets $R_i'$ are obtained. Section~\ref{subsec:reductions} describes fixed reduction methods, which construct $R_i'$ without using any information about the vehicles in the current episode. Section~\ref{subsec:topology_construction} introduces our algorithm, which constructs $R_i'$ from per-flow segment scores provided by a ranking oracle. Section~\ref{subsec:gnn_ranking} describes the proposed GNN approach for predicting the segment scores. Finally, Section~\ref{subsec:route_recovery} explains how a segment-flow solution is converted into individual vehicle routes. Figure~\ref{fig:schemes} illustrates the fixed reductions in panel~(a) and our proposed approach in panel~(b).

\subsection{Fixed reduction methods}
\label{subsec:reductions}

The fixed reduction scheme is illustrated in Figure~\ref{fig:schemes}(a). For each OD pair, the procedure generates a set of routes whose combined segments define the allowed set $R_i'$ -- once constructed, these sets remain unchanged across episodes. The approaches presented below represent different levels of pruning aggressiveness, from no pruning on the full graph to retaining only one shortest path for each OD pair.

\paragraph{Full graph (FG)}
FG applies no network reduction and sets $R_i'=R_i$ for every flow. The MILP is therefore solved on the complete road network, retaining all routing variables of the original formulation without prior segment removal. This yields the largest and most computationally demanding model among the evaluated approaches. We therefore use FG as the reference for measuring the objective quality of the reduced approaches.

\paragraph{Bounded Routes}
For each OD pair, the BR approach~\citep{svadlenka2023_1} precomputes a large pool of candidate routes. An A$^\ast$ search enumerates routes whose length lies within a factor $c>1$ of the shortest route, using the shortest-path length to the destination as the heuristic while respecting the allowed turns at each expansion. Every segment appearing in at least one enumerated route is then retained, and the union of these segments for the OD pair defines $R_i'$. BR thus discards segments absent from all selected routes while preserving several routing alternatives for each OD pair.

\paragraph{Smart Routes}
The SR approach~\citep{svadlenka2023_1} starts from the BR route pool and removes near-duplicate routes. DBSCAN clusters all available routes based on the Jaccard distance between their segment sets. Within each cluster, SR retains the shortest route, while routes classified as noise are likewise preserved as distinct alternatives. The corresponding $R_i'$ is then the union of the segments of the selected routes. Thus, SR preserves structurally distinct alternatives while eliminating routes that traverse largely overlapping segment sets, typically yielding a smaller flow-specific segment set than BR.

\paragraph{Shortest Path (SP)}
The SP approach retains only the shortest route satisfying the turn restrictions for each OD pair. Thus, SP represents the most aggressive fixed reduction considered in the evaluation, eliminating all alternative routing choices for each OD pair. Unlike BR and SR, it requires neither route enumeration nor clustering. As the extreme case of network reduction, SP provides a useful baseline for assessing the trade-off between model size and the loss of routing flexibility. In particular, comparison with less restrictive reductions indicates how much solution quality is affected when each OD pair is confined to a single predefined route.

We emphasize that the fixed approaches described above ignore the information about the vehicles in the current instance, and all episodes use the same flow-specific segment sets. Our approach, introduced next, instead adapts $R_i'$ to each CTR instance using both the information about the road network and the OD flows in the current episode.

\begin{algorithm}[!tp]
\caption{Ranking-guided network reduction}
\label{alg:ranking_reduction}
\begin{algorithmic}[1]
\Require CTR instance $I=(G,T,\{(o^i,d^i,n^i,R_i)\}_{i=1}^{N_F})$, per-flow segment scores $\{q_r^i\}$, rank cutoff $K_{\mathrm r}$, detour factor $\lambda$
\Ensure Flow-specific segment sets $\{R_i'\}_{i=1}^{N_F}$
\Statex \Comment{In Step 2, $\varnothing$ means no route passing the checks was obtained.}
\ForAll{flows $i$}
\Statex \textbf{Step 1: Initializing the route pool}
\State $p_0^i \gets \operatorname{ShortestODPath}(R_i,T,o^i,d^i;\ell)$
\State $\mathcal{P}_i \gets \{p_0^i\}$
\Statex \textbf{Step 2: Constructing OD routes using segment rankings}
\Statex \Comment{Phase 1: Segment selection}
\State $d_\tau^i \gets \operatorname{ShortestODCost}(R_i,T,o^i,d^i;\tau)$
\State $\mathcal{S}_i \gets \varnothing$
\ForAll{$r\in\operatorname{TopK}_{K_{\mathrm r}}(q^i|_{R_i})$ in ranking order}
\State $\hat{p}_r^i \gets \operatorname{BuildODRouteThrough}(r,R_i,T;\tau,\lambda d_\tau^i)$
\If{$\hat{p}_r^i\neq\varnothing$}
\State $\mathcal{S}_i \gets \mathcal{S}_i\cup\{r\}$
\EndIf
\EndFor
\State $\widehat{\mathcal{P}}_i \gets \{\hat{p}_r^i:r\in\mathcal{S}_i\}$
\Statex \Comment{Phase 2: Route construction}
\State $U_i \gets p_0^i$; $\rho^i \gets \operatorname{PercentileRanks}(q^i|_{R_i})$
\ForAll{$r\in\mathcal{S}_i$ in increasing order of $\rho_r^i$}
\If{$r\notin U_i$}
\State $p\gets\operatorname{RankedODPathThrough}(r,R_i,T,\rho^i;\tau,\lambda d_\tau^i)$
\If{$p=\varnothing$}
\State $p\gets\hat{p}_r^i$
\EndIf
\State $\mathcal{P}_i\gets\mathcal{P}_i\cup\{p\}$
\State $U_i\gets U_i\cup p$
\EndIf
\EndFor
\Statex \textbf{Step 3: Constructing the flow-specific segment set}
\State $E_i\gets\operatorname{ExpandOneTurn}(U_i,R_i,T)$
\State $R_i'\gets\operatorname{PruneToCompleteOD}(E_i,o^i,d^i,T)$
\EndFor
\State \Return $\{R_i'\}_{i=1}^{N_F}$
\end{algorithmic}
\end{algorithm}

\subsection{Ranking-guided network reduction}
\label{subsec:topology_construction}

The proposed approach is illustrated in Figure~\ref{fig:schemes}(b) and summarized in Algorithm~\ref{alg:ranking_reduction}. For each OD flow $i$, a ranking oracle assigns a score $q_r^i\in\mathbb{R}$ to every segment $r$, taking into account both the road network and the OD flow information in the current episode. The resulting scores therefore make the reduction specific to the current CTR episode. Given these scores, the procedure selects a limited set of ranked segments and constructs complete OD routes containing them. The union of these routes is then expanded locally to retain nearby routing alternatives. Once the procedure is finished, the original segment-flow MILP is solved on the union of all retained segments, with each flow $i$ restricted to its corresponding segment set $R_i'$.

\paragraph{Step 1: Initializing the route pool}

For each OD flow $i$, we first initialize the pool of routes $\mathcal{P}_i$ with the shortest (by length) path which satisfies the turn restrictions. We compute it using Dijkstra's algorithm, tracking both the current and previous junctions to enforce the turn restrictions. This route is retained regardless of the ranking, ensuring that at least one complete OD route remains available even if no additional ones are constructed.

\paragraph{Step 2: Constructing OD routes using segment rankings}

We start by considering the first $K_{\mathrm r}$ segments in the ranking and perform a further selection among them. For each candidate, we search for a shortest (by estimated travel time) OD path that passes through this segment. We use forward and reverse Dijkstra searches from the origin and destination, respectively, reusing their results for all candidates of the same OD pair. A segment is selected only when we find a path that respects the turn restrictions, contains no repeated junctions, and has travel time at most $\lambda d_\tau^i$. Here, $d_\tau^i$ is the shortest estimated travel time between the origin and destination without requiring the route to pass through the segment. Only the segments that do not fail this check proceed to the second phase. The routes found are stored separately in a pool $\widehat{\mathcal{P}}_i$, but are not yet added to $\mathcal{P}_i$.

In the second phase, we construct routes through the selected segments, starting with those ranked highest. Segments that are already included in at least one route of $\mathcal{P}_i$ are skipped. We first convert the ranking positions of the segments into values $\rho_r^i\in[0,1]$: the highest-ranked segment receives $0$, and the lowest-ranked segment receives $1$. Using these values as costs, we connect the origin to the start of the selected segment and its end to the destination through minimum-cost routes. Together with the selected segment, these form a complete OD route. If this route passes the same checks as in the first phase, we add it to $\mathcal{P}_i$; otherwise, we add the route through this segment stored in $\widehat{\mathcal{P}}_i$ during the first phase.

\paragraph{Step 3: Constructing the flow-specific segment set}

Once the pool of routes is constructed, we combine all segments used by these routes into a set $U_i$. We then expand this set by adding neighbouring segments that can directly precede or follow at least one segment in $U_i$ without violating the turn restrictions. We denote the resulting set by $E_i$. We then prune $E_i$, removing segments that are unreachable from the origin or from which the destination cannot be reached within this network. The remaining segments form the final set $R_i'$ available for routing flow $i$.

\paragraph{Optimization on the reduced network}

Once the flow-specific segment sets $R_i'$ are obtained, their union $R'=\bigcup_i R_i'$ defines the reduced segment set of the original segment-flow MILP, with each flow $i$ restricted to its corresponding $R_i'$. Variables $x_r^i$ are therefore created only for segments $r\in R_i'$, while the flow conservation and turn restrictions retain their original form. All other parts of the formulation remain unchanged. We note that the solver may combine segments in $R_i'$ into any routing allowed by the original constraints so the search space is not restricted to the routes in $\mathcal{P}_i$.

We remark that the reduction procedure does not depend on how the segment rankings are obtained. In this work, we propose to generate them using the GNN described in the following subsection.

\subsection{GNN-based segment ranking}
\label{subsec:gnn_ranking}

To obtain the segment rankings, we use a GNN that processes the full MILP instance before network reduction. The model is trained on FG solutions, where the number of vehicles on each segment determines the segment ordering for the corresponding flow (segments used by more vehicles rank higher). We next describe the construction and training of the GNN in detail.

\begin{figure}[!t]
\centering
\resizebox{\linewidth}{!}{%
\begin{tikzpicture}[
>=Latex,
fbox/.style={draw, rounded corners, align=center, font=\scriptsize,
inner sep=3pt, text width=32mm, minimum height=9mm},
obox/.style={draw, rounded corners, align=center, font=\scriptsize,
inner sep=3pt, text width=27mm, minimum height=9mm, fill=gray!8},
gnode/.style={circle, draw, minimum size=5.5mm, inner sep=0pt, fill=gray!15},
onode/.style={circle, draw, minimum size=5.8mm, inner sep=0pt, fill=gray!30},
gedge/.style={gray!50, line width=0.35pt},
mp/.style={violet!75, line width=1.0pt,
{Latex[length=1.6mm]}-{Latex[length=1.6mm]}},
flow/.style={-{Latex[length=2.1mm]}, line width=0.85pt},
hdr/.style={font=\scriptsize\itshape},
dot/.style={circle, draw=none, inner sep=0pt, minimum size=3.2pt},
]

\node[font=\footnotesize\bfseries] (features_title) at (0, 2.15) {node features};
\node[fbox, fill=blue!12]   (fmilp) at (0, 1.30) {\textbf{MILP}\\variable type, bounds, coefficients};
\node[fbox, fill=green!16]  (froad) at (0, 0.00) {\textbf{road network}\\length, speed, distances\\and turns};
\node[fbox, fill=orange!22] (fdem)  at (0,-1.30) {\textbf{OD flows}\\$n^i$, origin, destination};

\foreach \y/\n in {1.02/v1, 0.34/v2, -0.34/v3, -1.02/v4}
\node[gnode] (\n) at (3.65,\y) {};
\foreach \y/\n in {0.72/c1, 0/c2, -0.72/c3}
\node[gnode] (\n) at (5.20,\y) {};
\node[onode] (o1) at (6.50,0) {};

\node[hdr] (hv) at (3.65,1.50) {variables};
\node[hdr] (hc) at (5.20,1.22) {constraints};
\node[hdr] (ho) at (6.50,-0.56) {objective};

\foreach \v/\c in {v1/c1,v1/c2,v2/c1,v2/c2,v3/c2,v3/c3,v4/c2,v4/c3}
\draw[gedge] (\v)--(\c);
\foreach \c in {c1,c2,c3}
\draw[gedge] (\c)--(o1);
\draw[gedge] (v1) to[bend left=12] (o1);
\draw[gedge] (v4) to[bend right=12] (o1);

\draw[mp] (v2)--(c2);
\draw[mp] (v3)--(c3);
\draw[mp] (c2)--(o1);

\coordinate (gpadtop) at (5.10,1.58);
\coordinate (gpadbot) at (5.10,-1.58);
\node[draw, dashed, rounded corners, inner sep=3mm,
fit=(hv)(hc)(ho)(v1)(v4)(c1)(c3)(o1)(gpadtop)(gpadbot)] (graph) {};
\node[font=\footnotesize\bfseries, anchor=south] at ([yshift=1mm]graph.north)
{GATv2 message passing (4 layers)};

\node[dot, fill=blue!65] at ([yshift=0.75cm]graph.west) {};
\node[dot, fill=green!45!black] at (graph.west) {};
\node[dot, fill=orange!85!black] at ([yshift=-0.75cm]graph.west) {};
\draw[mp] ([xshift=4mm,yshift=3mm]graph.south west) -- ++(0.70,0);
\node[font=\scriptsize, anchor=west] at ([xshift=13mm,yshift=3mm]graph.south west)
{message passing};

\draw[->, blue!65, line width=0.8pt]         (fmilp.east) -- ([yshift=0.75cm]graph.west);
\draw[->, green!45!black, line width=0.8pt]  (froad.east) -- (graph.west);
\draw[->, orange!85!black, line width=0.8pt] (fdem.east)  -- ([yshift=-0.75cm]graph.west);

\node[obox] (scores) at (9.75, 0.68) {raw segment scores\\$x_r^i \mapsto q_r^i$};
\node[obox] (heat)   at (9.75,-0.82) {per-flow ranking for\\network reduction};
\draw[flow] ([yshift=0.40cm]graph.east) -- (scores.west);
\draw[flow] (scores) -- (heat);

\end{tikzpicture}%
}
\caption{GNN segment ranking model. The current CTR MILP instance is represented by variable, constraint, and objective nodes connected through six directed relation types. Node features combine MILP (\textcolor{blue!65}{blue}), road network (\textcolor{green!45!black}{green}), and OD-flow (\textcolor{orange!85!black}{orange}) information. Four GATv2 layers perform message passing (purple), after which a readout head assigns a raw score $q_r^i$ to each routing variable $x_r^i$. These scores are ranked separately within each OD flow and passed to the network reduction procedure.}
\label{fig:gnn_encoder}
\end{figure}
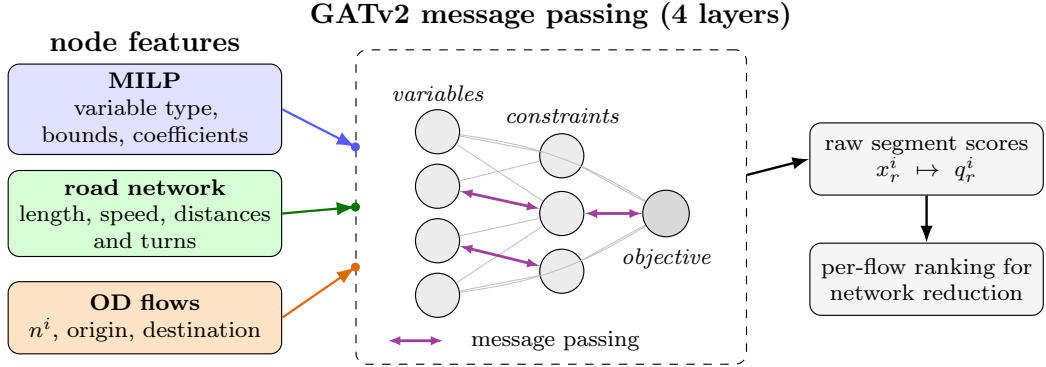

\paragraph{Representation of the problem instance}

The full MILP instance is represented as the heterogeneous graph shown in Figure~\ref{fig:gnn_encoder}. Following~\citet{Gasse2019}, each MILP variable and constraint is represented by a separate node. We additionally introduce a single objective node, resulting in three node types: variables, constraints, and objective. A variable node and a constraint node are connected if the variable has a nonzero coefficient in that constraint, and the coefficient is used as the edge attribute. Every variable and constraint node is also connected to the objective node; variable-objective edges carry the corresponding linear objective coefficients. All three connection types are represented in both directions, resulting in six directed relation types. Since the graph size is not fixed, the same GNN can process instances with different numbers of OD flows.

\paragraph{Node features}

The graph is augmented with node features that describe the current CTR instance from three perspectives: its MILP representation, fixed characteristics of the transportation network, and the OD flows in the current episode. The complete set of features for each node type is reported in Table~\ref{tab:gnn_features}. MILP-related features encode the variables and constraints and their role in the formulation, using information such as variable bounds, linear objective coefficients, constraint coefficients, and right-hand sides. Network features capture the physical and routing properties of individual segments, including travel characteristics, capacity, shortest-path position, and allowed turns. OD-flow features reflect the episode-specific use of the network, including flow sizes, flow-size-to-capacity ratios, and the sizes of other flows competing for the same segments.

\begin{table}[!t]
\centering
\caption{Node features used in the GNN representation.}
\label{tab:gnn_features}
\begin{tabular}{@{}l p{0.72\linewidth}@{}}
\toprule
Node type & Features \\
\midrule
Variable (77) & variable type, linear objective coefficient, and bounds; routing, network, and OD-flow features for $x_r^i$; turn context for $f_{a,b}^i$; segment features for $s_r$, $y_r$, and $z_r^k$, with additional tier-specific features for $z_r^k$ \\
Constraint (44) & constraint type, right-hand side, and sense; flow and OD context, segment and OD-flow features, congestion-tier information, and turn connectivity according to the constraint type \\
Objective (1) & a single bias feature \\
\bottomrule
\end{tabular}
\end{table}

\paragraph{Message passing and training}

The features of each node type are first projected into a common hidden space. Four message-passing layers then apply a separate GATv2 convolution~\citep{Brody2022} to each of the six directed relation types, using the corresponding edge attributes. The resulting messages are aggregated and combined with the previous node representation through a residual connection, followed by layer normalization, LeakyReLU, and dropout. A two-layer readout assigns a scalar score to every variable node, while only scores associated with routing variables $x_r^i$ are retained as $q_r^i$. The model is trained using FG solutions, with normalized segment usage $\bar{x}_r^i=x_r^i/n^i$ defining the target ordering within each flow. For a sampled pair $(r,s)$ with $\bar{x}_r^i>\bar{x}_s^i$, the loss is $\mathcal{L}_{i,r,s}=\sqrt{\bar{x}_r^i-\bar{x}_s^i}\,\operatorname{softplus}(q_s^i-q_r^i)$, which penalizes incorrect ordering and gives more weight to larger differences in segment usage.

\begin{algorithm}[!tp]
\caption{Route recovery with turn restrictions}
\label{alg:route_recovery}
\begin{algorithmic}[1]
\Require MILP solution $(x,f)$
\Ensure Vehicle routes $\{\mathcal{Q}_i\}_{i=1}^{N_F}$
\State $\mathcal{Q}_i\gets\varnothing$ for all $i$
\ForAll{flows $i$ with $n^i>0$}
\ForAll{$v\in V_T\setminus\{o^i,d^i\}$}
\If{some $f_{a,b}^i$, $(a,b)\in T_v$, is fractional}
\State $(f_{a,b}^i)_{(a,b)\in T_v}\gets\operatorname{MaximumFlow}(x^i,T_v)$
\EndIf
\EndFor
\State $x_{\mathrm{res}}^i\gets x^i$; $f_{\mathrm{res}}^i\gets f^i$
\Statex \Comment{DFS uses positive residual flows and never repeats a segment.}
\Statex \Comment{Backtrack over routes and vehicle counts if needed.}
\While{$|\mathcal{Q}_i|<n^i$}
\State $p\gets\operatorname{DFSRoute}(i,x_{\mathrm{res}}^i,f_{\mathrm{res}}^i)$
\State Choose $1\le h\le n^i-|\mathcal{Q}_i|$ supported by the residual flows along $p$
\State Add $h$ copies of $p$ to $\mathcal{Q}_i$; subtract $h$ from $x_{\mathrm{res}}^i$ and $f_{\mathrm{res}}^i$ along $p$
\EndWhile
\State Discard any residual circulations
\EndFor
\State \Return $\{\mathcal{Q}_i\}_{i=1}^{N_F}$
\end{algorithmic}
\end{algorithm}

\subsection{Route recovery}
\label{subsec:route_recovery}

Any solution returned by the CTR MILP must still be converted into individual vehicle routes. \citet{Nyporko2025} extract the solution using a DFS-based algorithm. However, the turn restrictions considered here require a different procedure. We therefore propose the route recovery procedure summarized in Algorithm~\ref{alg:route_recovery}.

As noted in Section~\ref{subsec:milp_formulation}, an integrality requirement is not imposed on the turn-flow variables $f_{a,b}^i$. The solver may return integer values for them, but it is not required to do so. If a given assignment is fractional at some restricted junction, we replace it with an integer one. For each flow $i$ and junction $v\in V_T\setminus\{o^i,d^i\}$, we formulate a maximum-flow problem: the nodes represent incoming and outgoing segments; we additionally introduce a source and a sink. We then connect the source to the incoming segments and the outgoing segments to the sink, adding arcs between segments only for allowed turns. The capacities of the arcs adjacent to the source and sink equal the corresponding $x_r^i$ values. Since all capacities are integer, the integrality property of maximum flow guarantees an integer solution. We remark that maximum flow can be solved in polynomial time.

Once an integer solution is available for all restricted junctions, we reconstruct the routes for each flow $i$. Initially, the residual flows equal the solution values $x_r^i$ on segments and $f_{a,b}^i$ on turns. A depth-first search finds a feasible OD route with no repeated segments using only positive residual flows. After assigning vehicles to the route, we subtract their number from the residual flows on its segments and turns and repeat the search using the updated values. The procedure continues until routes have been assigned to all $n^i$ vehicles. Any remaining flow forms circulations that carry no vehicles between the origin and the destination and is discarded.

\section{Experimental Evaluation}
\label{sec:experiments}

This section evaluates the proposed approach from four complementary perspectives. We first study how the considered CTR MILP scales with the number of OD flows and identify the settings in which it becomes difficult to solve within the operational budget, while additional computation can still improve the solution quality (Section~\ref{subsec:results_scalability}). Next, we assess how accurately the GNN ranks segments (Section~\ref{subsec:results_gnn}) and then compare the proposed approach with the other methods in terms of objective quality and runtime (Section~\ref{subsec:results_optimization}). Finally, we replay the obtained routes in a microscopic traffic simulator to assess how differences in the optimization translate into traffic outcomes (Section~\ref{subsec:results_simulation}).

\subsection{Experimental setup}
\label{subsec:setup}

\paragraph{Road networks}
The experiments are carried out on four real urban road networks -- Prague-Dejvice, London, New York, and Sydney -- spanning a range of sizes and topologies.  Table~\ref{tab:cities} reports their number of junctions, directed segments, and junctions with turn restrictions, together with the total segment length. The same rule is used for determining congestion-dependent segment costs in all four networks. Similar to \citet{Ekstrom2012} and \citet{Angelelli2021}, who use piecewise-linear approximations of congestion-dependent travel times, we represent each segment cost by ten tiers with quadratic threshold offsets and a fourth-power cost progression. To define its segment-specific thresholds, let $m_r$ denote the number of lanes represented by segment $r$, and let $\ell_{rj}$ be the length of lane $j$. Using an average vehicle spacing of $5$ metres, we define the nominal segment capacity as $\kappa_r=\sum_{j=1}^{m_r}\lceil\ell_{rj}/5\rceil$. The congestion thresholds are set relative to this capacity. The first nine tiers have finite upper thresholds $\theta_r^k=\kappa_r+(k-1)^2$, $k=1,\ldots,9$, while the tenth tier is unbounded. We next define the cost associated with each tier. The base cost is the expected traversal time of a lightly loaded segment. This time combines the travel time along the segment with the expected traffic-light delay at its downstream junction. We denote it by $\bar c_r$ and set $c_r^1=\max\{1,\lfloor\bar c_r\rfloor\}$. The remaining finite-tier costs are $c_r^k=\max\{c_r^{k-1}+1,\lfloor c_r^1+(5000-c_r^1)((k-1)/9)^4+1/2\rfloor\}$, $k=2,\ldots,9$, with $c_r^{10}=5000$.

\begin{table}[!t]
\centering
\caption{Topology of the four test cities.}
\label{tab:cities}
\begin{tabular}{l r r r r}
\toprule
City & $|V|$ & $|R|$ & $|T|$ & Length (km) \\
\midrule
Dejvice & 95 & 183 & 5 & 15.1 \\
London & 308 & 645 & 65 & 40.7 \\
New York & 212 & 453 & 9 & 51.8 \\
Sydney & 441 & 1058 & 20 & 106.4 \\
\bottomrule
\end{tabular}
\end{table}

\paragraph{Evaluation scenarios}

We retain the scenario structure used in~\citet{Nyporko2025} and extend it to four cities and a wider range of OD flows. Each scenario reflects one hour of peak-period traffic and consists of $120$ episodes of length $\Delta=30$ seconds. Each episode contains $N_F$ OD flows. How the number of vehicles in each flow evolves over the hour depends on the traffic regime. We consider two traffic regimes for each city: \emph{static} and \emph{increasing}. In the static regime, each flow independently follows either a constant or an increasing traffic pattern with equal probability; in the increasing regime, all flows follow the increasing traffic pattern. Both patterns are described below.

For a flow with a constant traffic pattern, the rate $\lambda_i$ is randomly selected from a discrete set spanning $0.10$--$0.67$ vehicles per second. With $\Delta=30$ seconds, this corresponds to $\widetilde n_e^i\in\{3,\ldots,20\}$ vehicles per episode, obtained by rounding $\lambda_i\Delta$. For city $c$, this number is scaled as $n_e^i=\max\{1,\lfloor\gamma_c\widetilde n_e^i+1/2\rfloor\}$, where $\gamma_c$ is the city-specific scaling factor. Since $\lambda_i$ is fixed, the number of vehicles remains constant throughout the one-hour scenario. 

For a flow with an increasing traffic pattern, one rate triple $\lambda_i^{(1)}<\lambda_i^{(2)}<\lambda_i^{(3)}$ is randomly selected and used over three consecutive $20$-minute periods, indexed by $s\in\{1,2,3\}$. These rates lie between $0.10$ and $0.44$ vehicles per second. For each episode $e$ in period $s(e)$, the number of vehicles $\widetilde n_e^i$ is sampled independently as $\widetilde n_e^i\sim\operatorname{Poisson}\!\left(\lambda_i^{(s(e))}\Delta\right)$. Thus, the number of vehicles varies across episodes, while its expected value increases in three stages over the hour. The same city-specific scaling is then applied to positive values of $\widetilde n_e^i$.

We use ten scenarios per experiment for every $(\text{city},\text{mode},N_F)$ combination; microscopic simulation reuses the final evaluation scenarios. The scaling factor $\gamma_c$ equals $0.80$, $0.72$, $0.67$, and $1.07$ for Prague-Dejvice, London, New York, and Sydney, respectively. In the optimization comparison, all methods receive a $10$-second operational solver budget. A separate $30$-second FG run provides a reference solution for each episode and allows us to assess the benefit of additional computation. This longer budget matches the episode length $\Delta$ but is not proposed for operational use, since preparing the optimization problem, recovering routes, and communicating them to vehicles also require time.

\paragraph{Compared methods}
The evaluation includes all methods introduced in Section~\ref{subsec:reductions} (FG, the fixed reduction methods SP, BR, and SR) as well as three ranking oracles that use the procedure described in Section~\ref{subsec:topology_construction}: GNN, linear programming (LP), and Random. The proposed GNN oracle ranks the segments of each flow using its predicted scores, as described in Section~\ref{subsec:gnn_ranking}. The LP oracle instead uses the fractional $x_r^i$ values obtained from the continuous relaxation of the FG, while Random uses a random segment permutation for each flow. All three oracles -- GNN, LP, and Random -- use the same $K_{\mathrm r}$ and $\lambda$, calibrated from the GNN training and validation data as described in Section~\ref{subsec:results_optimization}.

To assess the reliability of conclusions about differences between methods, we perform a statistical analysis of the obtained results. For each evaluated metric and traffic regime, we compare GNN with the best of the other methods using the All cities results, pairing their outcomes on the same scenarios. We construct $95\%$ confidence intervals using bootstrap and test the significance of differences using a two-sided paired randomization test. Since the best other method is selected from several alternatives rather than specified in advance, we apply Holm's correction to the $p$-values to reduce the risk of falsely declaring differences significant.

\paragraph{Implementation details}
The implementation is done in Python~3.13 and uses Gurobi~13.0.2, PyTorch Geometric, and SUMO~v1.27.1~\citep{SUMO2018}. The code and data are available in a public repository.\footnote{The repository link will be provided after review.} BR uses a pool of at most $100$ routes per OD pair and $c=1.3$; SR applies DBSCAN with $\varepsilon=0.3$ and a minimum of two samples. The GNN uses a hidden dimension of $128$, four attention heads, and a dropout rate of $0.1$. All experiments run on a NixOS server with an AMD EPYC~9124 processor ($64$ cores), $384$\,GB of RAM, and one NVIDIA L40S GPU with $48$\,GB of memory.

\subsection{Scalability of the MILP formulation}
\label{subsec:results_scalability}

This series of experiments studies how the considered CTR MILP scales as the number of OD flows increases. We are interested in the intermediate range of OD flows where the operational $10$-second budget is often insufficient to approach an optimum, but extending the solver budget to $30$ seconds still leads to substantially better results. This is the range where the solver can benefit the most from the proposed learned reduction: the problem is already nontrivial, so the solver cannot handle it within the operational budget; however, the improvement under the longer budget indicates that high-quality solutions remain attainable, motivating the use of these instances for GNN training. To identify the targeted intermediate range for each road network and traffic regime, we vary the number of origin-destination pairs, considering $N_F\in\{5,10,15,20,25,30,35,40,45,50\}$.

We use two metrics to characterize the solver's performance under both the operational and extended time budgets. The first is the fraction of episodes for which FG obtains a feasible solution with a primal--dual gap of at most $1\%$. The gap is defined as $g_{\mathrm{PD}}=(z_{\mathrm{best}}-z_{\mathrm{bound}})/|z_{\mathrm{best}}|$, where $z_{\mathrm{best}}$ is the best feasible objective value found and $z_{\mathrm{bound}}$ is the proven lower bound. The second metric we use is the fraction of episodes for which FG obtains any feasible solution, regardless of its gap. Figure~\ref{fig:scalability} demonstrates the results.

\definecolor{expbudget0}{RGB}{232,91,52}
\definecolor{expbudget1}{RGB}{28,139,210}
\definecolor{expselection}{RGB}{218,235,211}
\definecolor{expselectionedge}{RGB}{117,149,108}
\begin{figure*}[!t]
\centering
{\scriptsize \tikz[baseline=-0.6ex]\draw[expbudget0,thick,mark=*] plot coordinates {(0,0) (0.55,0)}; 10 s\qquad \tikz[baseline=-0.6ex]\draw[expbudget1,thick,mark=square*] plot coordinates {(0,0) (0.55,0)}; 30 s\qquad \tikz[baseline=-0.6ex]\draw[black,thick] (0,0)--(0.55,0); gap $\leq 1\%$\qquad \tikz[baseline=-0.6ex]\draw[black,thick,dashed] (0,0)--(0.55,0); feasible solution}\par\smallskip
\begin{minipage}{0.49\linewidth}
\centering
\begin{tikzpicture}
\begin{axis}[
  width=0.98\linewidth, height=4.25cm,
  title={Prague-Dejvice -- Static},
  title style={font=\small},
  xmin=4.5, xmax=50.5,
  xtick={5,10,15,20,25,30,35,40,45,50},
  xlabel={},
  ylabel={Episodes (\%)},
  ymajorgrids=true,
  major grid style={line width=0.1pt, gray!30},
  tick label style={font=\scriptsize},
  label style={font=\footnotesize},
  unbounded coords=jump,
  ymin=-5, ymax=105,
  ytick={0,20,40,60,80,100},
]
\addplot[draw=none,fill=expselection,fill opacity=0.32,mark=none,forget plot]
  coordinates {
    (28,-5) (32,-5) (32,105) (28,105)
  } \closedcycle;
\addplot[expselectionedge,line width=0.45pt,mark=none,forget plot]
  coordinates {(28,-5) (28,105)};
\addplot[expselectionedge,line width=0.45pt,mark=none,forget plot]
  coordinates {(32,-5) (32,105)};
\addplot+[expbudget0, thick, mark=*, mark size=1.25pt, mark options={draw=expbudget0,fill=expbudget0}]
  coordinates {
    (5, 100)
    (10, 100)
    (15, 100)
    (20, 99.916667)
    (25, 89)
    (30, 44.666667)
    (35, 16.916667)
    (40, 4.3333333)
    (45, 0.16666667)
    (50, 0)
  };
\addplot+[expbudget0, thick, dashed, mark=*, mark size=1.25pt, mark options={draw=expbudget0,fill=expbudget0}]
  coordinates {
    (5, 100)
    (10, 100)
    (15, 100)
    (20, 100)
    (25, 100)
    (30, 100)
    (35, 100)
    (40, 100)
    (45, 100)
    (50, 100)
  };
\addplot+[expbudget1, thick, mark=square*, mark size=1.25pt, mark options={draw=expbudget1,fill=expbudget1}]
  coordinates {
    (5, 100)
    (10, 100)
    (15, 100)
    (20, 100)
    (25, 100)
    (30, 97.916667)
    (35, 71.666667)
    (40, 40.083333)
    (45, 6.9166667)
    (50, 0)
  };
\addplot+[expbudget1, thick, dashed, mark=square*, mark size=1.25pt, mark options={draw=expbudget1,fill=expbudget1}]
  coordinates {
    (5, 100)
    (10, 100)
    (15, 100)
    (20, 100)
    (25, 100)
    (30, 100)
    (35, 100)
    (40, 100)
    (45, 100)
    (50, 100)
  };
\end{axis}
\end{tikzpicture}
\end{minipage}
\hfill
\begin{minipage}{0.49\linewidth}
\centering
\begin{tikzpicture}
\begin{axis}[
  width=0.98\linewidth, height=4.25cm,
  title={Prague-Dejvice -- Increasing},
  title style={font=\small},
  xmin=4.5, xmax=50.5,
  xtick={5,10,15,20,25,30,35,40,45,50},
  xlabel={},
  ylabel={},
  ymajorgrids=true,
  major grid style={line width=0.1pt, gray!30},
  tick label style={font=\scriptsize},
  label style={font=\footnotesize},
  unbounded coords=jump,
  ymin=-5, ymax=105,
  ytick={0,20,40,60,80,100},
]
\addplot[draw=none,fill=expselection,fill opacity=0.32,mark=none,forget plot]
  coordinates {
    (31,-5) (35,-5) (35,105) (31,105)
  } \closedcycle;
\addplot[expselectionedge,line width=0.45pt,mark=none,forget plot]
  coordinates {(31,-5) (31,105)};
\addplot[expselectionedge,line width=0.45pt,mark=none,forget plot]
  coordinates {(35,-5) (35,105)};
\addplot+[expbudget0, thick, mark=*, mark size=1.25pt, mark options={draw=expbudget0,fill=expbudget0}]
  coordinates {
    (5, 100)
    (10, 100)
    (15, 100)
    (20, 99)
    (25, 85.25)
    (30, 65.083333)
    (35, 52.5)
    (40, 34.083333)
    (45, 15.833333)
    (50, 3.1666667)
  };
\addplot+[expbudget0, thick, dashed, mark=*, mark size=1.25pt, mark options={draw=expbudget0,fill=expbudget0}]
  coordinates {
    (5, 100)
    (10, 100)
    (15, 100)
    (20, 100)
    (25, 100)
    (30, 100)
    (35, 100)
    (40, 100)
    (45, 99.75)
    (50, 100)
  };
\addplot+[expbudget1, thick, mark=square*, mark size=1.25pt, mark options={draw=expbudget1,fill=expbudget1}]
  coordinates {
    (5, 100)
    (10, 100)
    (15, 100)
    (20, 100)
    (25, 99.916667)
    (30, 89.5)
    (35, 71.333333)
    (40, 61.583333)
    (45, 45.666667)
    (50, 26.333333)
  };
\addplot+[expbudget1, thick, dashed, mark=square*, mark size=1.25pt, mark options={draw=expbudget1,fill=expbudget1}]
  coordinates {
    (5, 100)
    (10, 100)
    (15, 100)
    (20, 100)
    (25, 100)
    (30, 100)
    (35, 100)
    (40, 100)
    (45, 100)
    (50, 100)
  };
\end{axis}
\end{tikzpicture}
\end{minipage}
\par\smallskip
\begin{minipage}{0.49\linewidth}
\centering
\begin{tikzpicture}
\begin{axis}[
  width=0.98\linewidth, height=4.25cm,
  title={London -- Static},
  title style={font=\small},
  xmin=4.5, xmax=50.5,
  xtick={5,10,15,20,25,30,35,40,45,50},
  xlabel={},
  ylabel={Episodes (\%)},
  ymajorgrids=true,
  major grid style={line width=0.1pt, gray!30},
  tick label style={font=\scriptsize},
  label style={font=\footnotesize},
  unbounded coords=jump,
  ymin=-5, ymax=105,
  ytick={0,20,40,60,80,100},
]
\addplot[draw=none,fill=expselection,fill opacity=0.32,mark=none,forget plot]
  coordinates {
    (7,-5) (11,-5) (11,105) (7,105)
  } \closedcycle;
\addplot[expselectionedge,line width=0.45pt,mark=none,forget plot]
  coordinates {(7,-5) (7,105)};
\addplot[expselectionedge,line width=0.45pt,mark=none,forget plot]
  coordinates {(11,-5) (11,105)};
\addplot+[expbudget0, thick, mark=*, mark size=1.25pt, mark options={draw=expbudget0,fill=expbudget0}]
  coordinates {
    (5, 100)
    (10, 57.25)
    (15, 6.75)
    (20, 0)
    (25, 0)
    (30, 0)
    (35, 0)
    (40, 0)
    (45, 0)
    (50, 0)
  };
\addplot+[expbudget0, thick, dashed, mark=*, mark size=1.25pt, mark options={draw=expbudget0,fill=expbudget0}]
  coordinates {
    (5, 100)
    (10, 100)
    (15, 100)
    (20, 47.416667)
    (25, 0.083333333)
    (30, 0)
    (35, 0)
    (40, 0)
    (45, 0)
    (50, 0)
  };
\addplot+[expbudget1, thick, mark=square*, mark size=1.25pt, mark options={draw=expbudget1,fill=expbudget1}]
  coordinates {
    (5, 100)
    (10, 97.083333)
    (15, 51.666667)
    (20, 1.6666667)
    (25, 0)
    (30, 0)
    (35, 0)
    (40, 0)
    (45, 0)
    (50, 0)
  };
\addplot+[expbudget1, thick, dashed, mark=square*, mark size=1.25pt, mark options={draw=expbudget1,fill=expbudget1}]
  coordinates {
    (5, 100)
    (10, 100)
    (15, 100)
    (20, 91.583333)
    (25, 52.5)
    (30, 45.083333)
    (35, 25.583333)
    (40, 14.833333)
    (45, 6.25)
    (50, 0.83333333)
  };
\end{axis}
\end{tikzpicture}
\end{minipage}
\hfill
\begin{minipage}{0.49\linewidth}
\centering
\begin{tikzpicture}
\begin{axis}[
  width=0.98\linewidth, height=4.25cm,
  title={London -- Increasing},
  title style={font=\small},
  xmin=4.5, xmax=50.5,
  xtick={5,10,15,20,25,30,35,40,45,50},
  xlabel={},
  ylabel={},
  ymajorgrids=true,
  major grid style={line width=0.1pt, gray!30},
  tick label style={font=\scriptsize},
  label style={font=\footnotesize},
  unbounded coords=jump,
  ymin=-5, ymax=105,
  ytick={0,20,40,60,80,100},
]
\addplot[draw=none,fill=expselection,fill opacity=0.32,mark=none,forget plot]
  coordinates {
    (10,-5) (14,-5) (14,105) (10,105)
  } \closedcycle;
\addplot[expselectionedge,line width=0.45pt,mark=none,forget plot]
  coordinates {(10,-5) (10,105)};
\addplot[expselectionedge,line width=0.45pt,mark=none,forget plot]
  coordinates {(14,-5) (14,105)};
\addplot+[expbudget0, thick, mark=*, mark size=1.25pt, mark options={draw=expbudget0,fill=expbudget0}]
  coordinates {
    (5, 100)
    (10, 73.5)
    (15, 31.666667)
    (20, 1.4166667)
    (25, 0)
    (30, 0)
    (35, 0)
    (40, 0)
    (45, 0)
    (50, 0)
  };
\addplot+[expbudget0, thick, dashed, mark=*, mark size=1.25pt, mark options={draw=expbudget0,fill=expbudget0}]
  coordinates {
    (5, 100)
    (10, 100)
    (15, 100)
    (20, 66.666667)
    (25, 1.1666667)
    (30, 0)
    (35, 0)
    (40, 0)
    (45, 0)
    (50, 0)
  };
\addplot+[expbudget1, thick, mark=square*, mark size=1.25pt, mark options={draw=expbudget1,fill=expbudget1}]
  coordinates {
    (5, 100)
    (10, 98.75)
    (15, 63.916667)
    (20, 23.083333)
    (25, 0.083333333)
    (30, 0)
    (35, 0)
    (40, 0)
    (45, 0)
    (50, 0)
  };
\addplot+[expbudget1, thick, dashed, mark=square*, mark size=1.25pt, mark options={draw=expbudget1,fill=expbudget1}]
  coordinates {
    (5, 100)
    (10, 100)
    (15, 100)
    (20, 91.416667)
    (25, 58.5)
    (30, 48.166667)
    (35, 35.583333)
    (40, 28.75)
    (45, 16.416667)
    (50, 3.6666667)
  };
\end{axis}
\end{tikzpicture}
\end{minipage}
\par\smallskip
\begin{minipage}{0.49\linewidth}
\centering
\begin{tikzpicture}
\begin{axis}[
  width=0.98\linewidth, height=4.25cm,
  title={New York -- Static},
  title style={font=\small},
  xmin=4.5, xmax=50.5,
  xtick={5,10,15,20,25,30,35,40,45,50},
  xlabel={},
  ylabel={Episodes (\%)},
  ymajorgrids=true,
  major grid style={line width=0.1pt, gray!30},
  tick label style={font=\scriptsize},
  label style={font=\footnotesize},
  unbounded coords=jump,
  ymin=-5, ymax=105,
  ytick={0,20,40,60,80,100},
]
\addplot[draw=none,fill=expselection,fill opacity=0.32,mark=none,forget plot]
  coordinates {
    (12,-5) (16,-5) (16,105) (12,105)
  } \closedcycle;
\addplot[expselectionedge,line width=0.45pt,mark=none,forget plot]
  coordinates {(12,-5) (12,105)};
\addplot[expselectionedge,line width=0.45pt,mark=none,forget plot]
  coordinates {(16,-5) (16,105)};
\addplot+[expbudget0, thick, mark=*, mark size=1.25pt, mark options={draw=expbudget0,fill=expbudget0}]
  coordinates {
    (5, 100)
    (10, 99.416667)
    (15, 2.0833333)
    (20, 0)
    (25, 0)
    (30, 0)
    (35, 0)
    (40, 0)
    (45, 0)
    (50, 0)
  };
\addplot+[expbudget0, thick, dashed, mark=*, mark size=1.25pt, mark options={draw=expbudget0,fill=expbudget0}]
  coordinates {
    (5, 100)
    (10, 100)
    (15, 89.75)
    (20, 11.083333)
    (25, 0.083333333)
    (30, 0)
    (35, 0)
    (40, 0)
    (45, 0)
    (50, 0)
  };
\addplot+[expbudget1, thick, mark=square*, mark size=1.25pt, mark options={draw=expbudget1,fill=expbudget1}]
  coordinates {
    (5, 100)
    (10, 100)
    (15, 97.833333)
    (20, 62.166667)
    (25, 17.25)
    (30, 0.25)
    (35, 0)
    (40, 0)
    (45, 0)
    (50, 0)
  };
\addplot+[expbudget1, thick, dashed, mark=square*, mark size=1.25pt, mark options={draw=expbudget1,fill=expbudget1}]
  coordinates {
    (5, 100)
    (10, 100)
    (15, 100)
    (20, 99.083333)
    (25, 96)
    (30, 72.833333)
    (35, 21.5)
    (40, 3.0833333)
    (45, 0.16666667)
    (50, 0)
  };
\end{axis}
\end{tikzpicture}
\end{minipage}
\hfill
\begin{minipage}{0.49\linewidth}
\centering
\begin{tikzpicture}
\begin{axis}[
  width=0.98\linewidth, height=4.25cm,
  title={New York -- Increasing},
  title style={font=\small},
  xmin=4.5, xmax=50.5,
  xtick={5,10,15,20,25,30,35,40,45,50},
  xlabel={},
  ylabel={},
  ymajorgrids=true,
  major grid style={line width=0.1pt, gray!30},
  tick label style={font=\scriptsize},
  label style={font=\footnotesize},
  unbounded coords=jump,
  ymin=-5, ymax=105,
  ytick={0,20,40,60,80,100},
]
\addplot[draw=none,fill=expselection,fill opacity=0.32,mark=none,forget plot]
  coordinates {
    (12,-5) (16,-5) (16,105) (12,105)
  } \closedcycle;
\addplot[expselectionedge,line width=0.45pt,mark=none,forget plot]
  coordinates {(12,-5) (12,105)};
\addplot[expselectionedge,line width=0.45pt,mark=none,forget plot]
  coordinates {(16,-5) (16,105)};
\addplot+[expbudget0, thick, mark=*, mark size=1.25pt, mark options={draw=expbudget0,fill=expbudget0}]
  coordinates {
    (5, 100)
    (10, 99.5)
    (15, 14.416667)
    (20, 0)
    (25, 0)
    (30, 0)
    (35, 0)
    (40, 0)
    (45, 0)
    (50, 0)
  };
\addplot+[expbudget0, thick, dashed, mark=*, mark size=1.25pt, mark options={draw=expbudget0,fill=expbudget0}]
  coordinates {
    (5, 100)
    (10, 100)
    (15, 80.333333)
    (20, 24.416667)
    (25, 2.5833333)
    (30, 0.25)
    (35, 0)
    (40, 0)
    (45, 0)
    (50, 0)
  };
\addplot+[expbudget1, thick, mark=square*, mark size=1.25pt, mark options={draw=expbudget1,fill=expbudget1}]
  coordinates {
    (5, 100)
    (10, 100)
    (15, 95.25)
    (20, 64)
    (25, 18.25)
    (30, 0)
    (35, 0)
    (40, 0)
    (45, 0)
    (50, 0)
  };
\addplot+[expbudget1, thick, dashed, mark=square*, mark size=1.25pt, mark options={draw=expbudget1,fill=expbudget1}]
  coordinates {
    (5, 100)
    (10, 100)
    (15, 100)
    (20, 99.083333)
    (25, 96.916667)
    (30, 46.166667)
    (35, 9)
    (40, 1)
    (45, 0.083333333)
    (50, 0)
  };
\end{axis}
\end{tikzpicture}
\end{minipage}
\par\smallskip
\begin{minipage}{0.49\linewidth}
\centering
\begin{tikzpicture}
\begin{axis}[
  width=0.98\linewidth, height=4.25cm,
  title={Sydney -- Static},
  title style={font=\small},
  xmin=4.5, xmax=50.5,
  xtick={5,10,15,20,25,30,35,40,45,50},
  xlabel={$N_F$},
  ylabel={Episodes (\%)},
  ymajorgrids=true,
  major grid style={line width=0.1pt, gray!30},
  tick label style={font=\scriptsize},
  label style={font=\footnotesize},
  unbounded coords=jump,
  ymin=-5, ymax=105,
  ytick={0,20,40,60,80,100},
]
\addplot[draw=none,fill=expselection,fill opacity=0.32,mark=none,forget plot]
  coordinates {
    (5,-5) (9,-5) (9,105) (5,105)
  } \closedcycle;
\addplot[expselectionedge,line width=0.45pt,mark=none,forget plot]
  coordinates {(5,-5) (5,105)};
\addplot[expselectionedge,line width=0.45pt,mark=none,forget plot]
  coordinates {(9,-5) (9,105)};
\addplot+[expbudget0, thick, mark=*, mark size=1.25pt, mark options={draw=expbudget0,fill=expbudget0}]
  coordinates {
    (5, 87.5)
    (10, 0.083333333)
    (15, 0)
    (20, 0)
    (25, 0)
    (30, 0)
    (35, 0)
    (40, 0)
    (45, 0)
    (50, 0)
  };
\addplot+[expbudget0, thick, dashed, mark=*, mark size=1.25pt, mark options={draw=expbudget0,fill=expbudget0}]
  coordinates {
    (5, 100)
    (10, 15.5)
    (15, 0.41666667)
    (20, 0)
    (25, 0)
    (30, 0)
    (35, 0)
    (40, 0)
    (45, 0)
    (50, 0)
  };
\addplot+[expbudget1, thick, mark=square*, mark size=1.25pt, mark options={draw=expbudget1,fill=expbudget1}]
  coordinates {
    (5, 99.833333)
    (10, 62.166667)
    (15, 1.9166667)
    (20, 0)
    (25, 0)
    (30, 0)
    (35, 0)
    (40, 0)
    (45, 0)
    (50, 0)
  };
\addplot+[expbudget1, thick, dashed, mark=square*, mark size=1.25pt, mark options={draw=expbudget1,fill=expbudget1}]
  coordinates {
    (5, 100)
    (10, 100)
    (15, 18.166667)
    (20, 0)
    (25, 0)
    (30, 0)
    (35, 0)
    (40, 0)
    (45, 0)
    (50, 0)
  };
\end{axis}
\end{tikzpicture}
\end{minipage}
\hfill
\begin{minipage}{0.49\linewidth}
\centering
\begin{tikzpicture}
\begin{axis}[
  width=0.98\linewidth, height=4.25cm,
  title={Sydney -- Increasing},
  title style={font=\small},
  xmin=4.5, xmax=50.5,
  xtick={5,10,15,20,25,30,35,40,45,50},
  xlabel={$N_F$},
  ylabel={},
  ymajorgrids=true,
  major grid style={line width=0.1pt, gray!30},
  tick label style={font=\scriptsize},
  label style={font=\footnotesize},
  unbounded coords=jump,
  ymin=-5, ymax=105,
  ytick={0,20,40,60,80,100},
]
\addplot[draw=none,fill=expselection,fill opacity=0.32,mark=none,forget plot]
  coordinates {
    (6,-5) (10,-5) (10,105) (6,105)
  } \closedcycle;
\addplot[expselectionedge,line width=0.45pt,mark=none,forget plot]
  coordinates {(6,-5) (6,105)};
\addplot[expselectionedge,line width=0.45pt,mark=none,forget plot]
  coordinates {(10,-5) (10,105)};
\addplot+[expbudget0, thick, mark=*, mark size=1.25pt, mark options={draw=expbudget0,fill=expbudget0}]
  coordinates {
    (5, 84.833333)
    (10, 22.5)
    (15, 0)
    (20, 0)
    (25, 0)
    (30, 0)
    (35, 0)
    (40, 0)
    (45, 0)
    (50, 0)
  };
\addplot+[expbudget0, thick, dashed, mark=*, mark size=1.25pt, mark options={draw=expbudget0,fill=expbudget0}]
  coordinates {
    (5, 100)
    (10, 49.833333)
    (15, 0.33333333)
    (20, 0)
    (25, 0)
    (30, 0)
    (35, 0)
    (40, 0)
    (45, 0)
    (50, 0)
  };
\addplot+[expbudget1, thick, mark=square*, mark size=1.25pt, mark options={draw=expbudget1,fill=expbudget1}]
  coordinates {
    (5, 99.75)
    (10, 73.083333)
    (15, 9.0833333)
    (20, 0)
    (25, 0)
    (30, 0)
    (35, 0)
    (40, 0)
    (45, 0)
    (50, 0)
  };
\addplot+[expbudget1, thick, dashed, mark=square*, mark size=1.25pt, mark options={draw=expbudget1,fill=expbudget1}]
  coordinates {
    (5, 100)
    (10, 100)
    (15, 36.833333)
    (20, 0)
    (25, 0)
    (30, 0)
    (35, 0)
    (40, 0)
    (45, 0)
    (50, 0)
  };
\end{axis}
\end{tikzpicture}
\end{minipage}
\par
\caption{%
FG scalability. Solid: gap $\leq 1\%$; dashed: feasible solution; shading: selected $N_F$.
}
\label{fig:scalability}
\end{figure*}
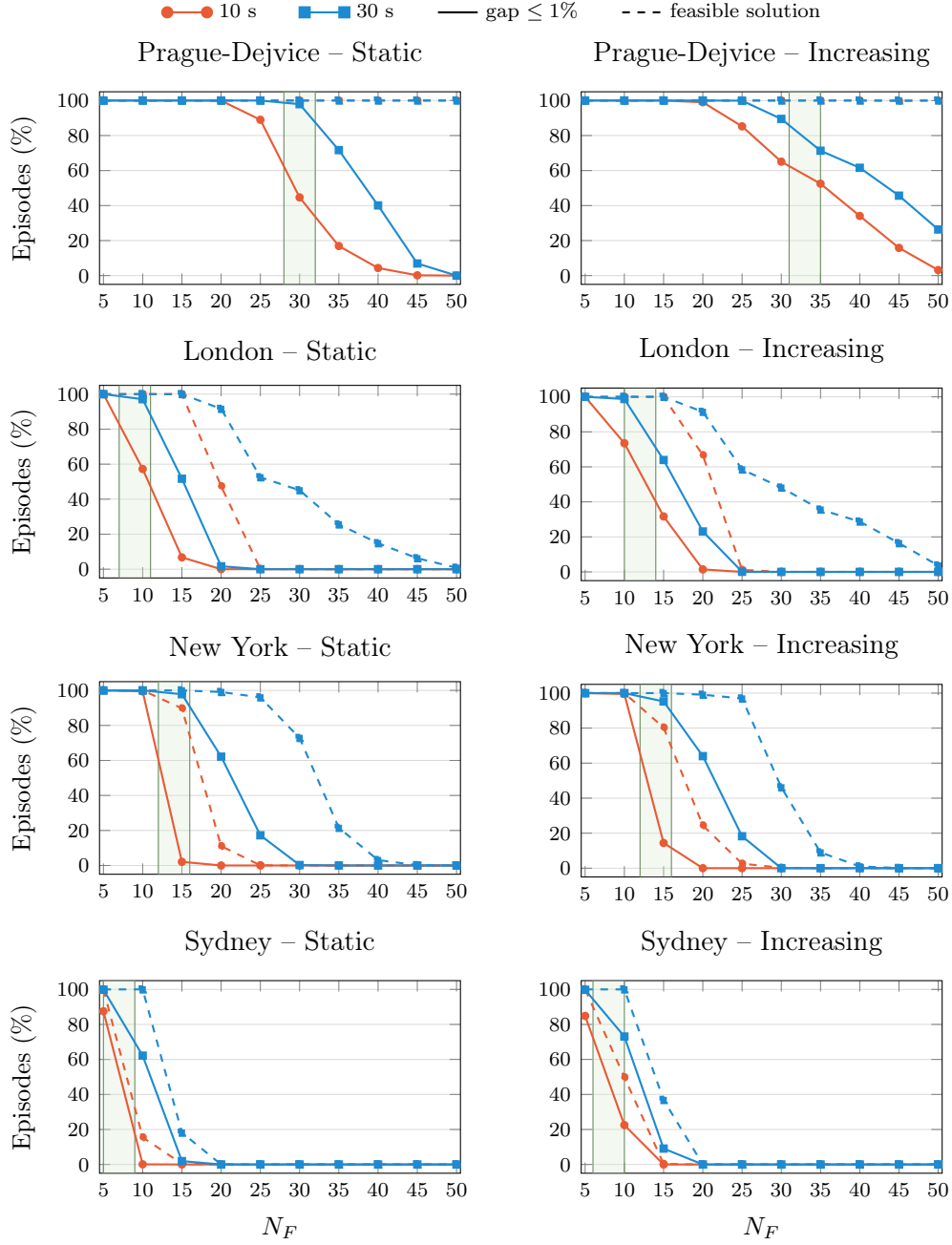

\paragraph{Discussion}
We first analyze the relation between the solid and dashed curves, which report, respectively, the fractions of episodes in which FG obtains a solution with a primal--dual gap of at most $1\%$ and any feasible solution. We make two observations. First, the values of $N_F$ at which the curves begin to decline broadly follow network size: Prague-Dejvice starts to decrease only at large $N_F$, London and New York do so in the middle range, while Sydney becomes computationally hard already at small $N_F$. Second, the distance between the solid and dashed curves shows whether the main difficulty lies in improving an already found solution or finding a feasible solution at all. In Prague-Dejvice, a feasible solution is found in almost every episode (a minimum of $99.75\%$) over the entire evaluated range, so the decrease of the solid curves mainly reflects the increasing difficulty of reaching a small primal-dual gap. London and New York show a two-stage trend: the solid curves decrease first while feasible solutions are still found in most episodes. For Sydney, the distance between the curves is the smallest, so finding a feasible solution becomes almost as difficult as obtaining a high-quality one, and feasibility itself quickly becomes computationally hard even for moderate numbers of OD flows.

We next identify the settings in which the problem is hard to tackle within the operational $10$-second budget, but extending the solver budget to $30$ seconds can still improve the solution. The shaded regions in Figure~\ref{fig:scalability} mark the corresponding ranges of five consecutive $N_F$ values selected for each city and traffic regime. We place each region around the point where the $10$-second solid curve starts to decline while the $30$-second curve remains substantially higher. This is the regime in which the proposed reduction can provide the clearest benefit: the full formulation is already difficult to solve within the operational budget, but the longer runs show that high-quality solutions remain attainable. The selected ranges are $28$--$32$ and $31$--$35$ for Prague-Dejvice, $7$--$11$ and $10$--$14$ for London, $12$--$16$ for both New York modes, and $5$--$9$ and $6$--$10$ for Sydney, with the static regime listed first. These city-specific ranges are used in all subsequent experiments.

\subsection{GNN training results}
\label{subsec:results_gnn}

This series of experiments demonstrates the ability of the proposed GNN to learn the flow-specific segment ordering. We train a separate GNN model (with the same architecture but different learned weights) for each city and traffic regime. Each model is trained jointly on the five corresponding $N_F$ values selected in the previous scalability experiment, using ten scenarios ($1200$ episodes) for each $N_F$ value. For every $N_F$, eight complete scenarios are used for training and two for validation. To obtain high-quality labels, we solve FG with a $300$-second time limit and retain every episode for which a feasible solution is found. All labelling runs produced a feasible solution, yielding $6000$ labelled MILP instances per model. The problem representation, target ordering, and training loss are described in Section~\ref{subsec:gnn_ranking}. Training uses Adam with learning rate $5\times10^{-3}$, weight decay $10^{-4}$, batch size $8$, and a cosine learning rate schedule for at most $50$ epochs. Early stopping is applied with a patience of five epochs.

We evaluate ranking quality using pairwise accuracy and top-$K$ recall. For each flow, we sample $128$ segment pairs with different target values and weight them as defined in Section~\ref{subsec:gnn_ranking}. A sampled pair with $\bar{x}_r^i>\bar{x}_s^i$ is correctly ordered when $q_r^i>q_s^i$. Pairwise accuracy is thus defined as the weighted fraction of correctly ordered pairs. The second metric, top-$K$ recall, considers the segments used by a flow $i$ in the labelled solution. It measures the fraction of such segments that appear among the $K$ highest-scoring segments for that flow. We report the mean top-$K$ recall across flows and validation episodes for $K\in\{10,20,50,100,200\}$. For model selection, validation pairwise accuracy is computed separately for each $N_F$, and the model with the highest mean across the five values is retained.

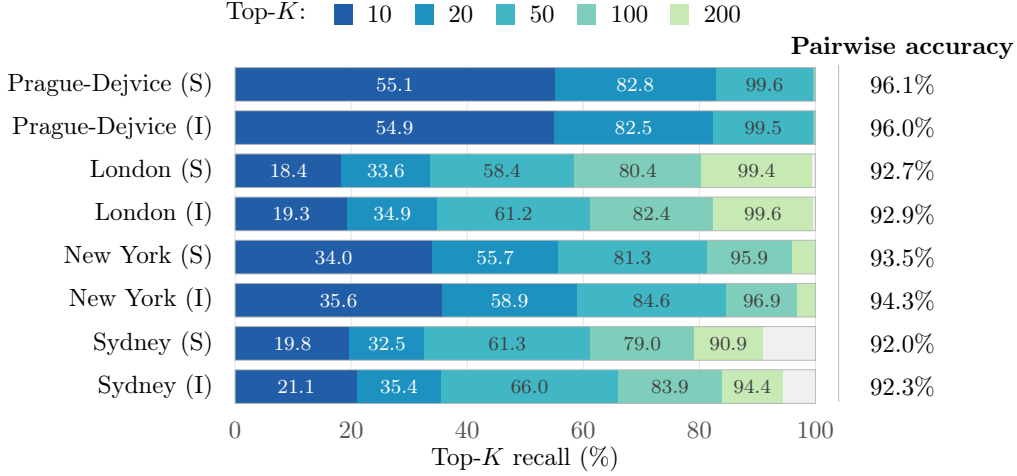
\begin{figure}[!t]
\centering
\definecolor{recallten}{RGB}{34,94,168}
\definecolor{recalltwenty}{RGB}{29,145,192}
\definecolor{recallfifty}{RGB}{65,182,196}
\definecolor{recallhundred}{RGB}{127,205,187}
\definecolor{recalltwohundred}{RGB}{199,233,180}
\definecolor{recallmissing}{RGB}{238,240,242}
\resizebox{\linewidth}{!}{%
\begin{tikzpicture}[x=0.82mm,y=6.1mm]
\begin{scope}[xshift=32mm]
\node[anchor=east,font=\footnotesize] at (14,9.35) {Top-$K$:};
\fill[recallten] (17,9.08) rectangle ++(3.2,0.48);
\node[anchor=west,font=\footnotesize] at (21.0,9.32) {10};
\fill[recalltwenty] (31,9.08) rectangle ++(3.2,0.48);
\node[anchor=west,font=\footnotesize] at (35.0,9.32) {20};
\fill[recallfifty] (45,9.08) rectangle ++(3.2,0.48);
\node[anchor=west,font=\footnotesize] at (49.0,9.32) {50};
\fill[recallhundred] (59,9.08) rectangle ++(3.2,0.48);
\node[anchor=west,font=\footnotesize] at (63.0,9.32) {100};
\fill[recalltwohundred] (75,9.08) rectangle ++(3.2,0.48);
\node[anchor=west,font=\footnotesize] at (79.0,9.32) {200};
\node[anchor=center,font=\footnotesize\bfseries] at (115,8.55) {Pairwise accuracy};
\draw[gray!55,line width=0.4pt] (104,0.35) -- (104,8.15);
\draw[gray!22,line width=0.25pt] (0,0.42) -- (0,8.12);
\node[anchor=north,font=\footnotesize,text=gray!65!black] at (0,0.12) {0};
\draw[gray!22,line width=0.25pt] (20,0.42) -- (20,8.12);
\node[anchor=north,font=\footnotesize,text=gray!65!black] at (20,0.12) {20};
\draw[gray!22,line width=0.25pt] (40,0.42) -- (40,8.12);
\node[anchor=north,font=\footnotesize,text=gray!65!black] at (40,0.12) {40};
\draw[gray!22,line width=0.25pt] (60,0.42) -- (60,8.12);
\node[anchor=north,font=\footnotesize,text=gray!65!black] at (60,0.12) {60};
\draw[gray!22,line width=0.25pt] (80,0.42) -- (80,8.12);
\node[anchor=north,font=\footnotesize,text=gray!65!black] at (80,0.12) {80};
\draw[gray!22,line width=0.25pt] (100,0.42) -- (100,8.12);
\node[anchor=north,font=\footnotesize,text=gray!65!black] at (100,0.12) {100};
\node[anchor=east,font=\footnotesize] at (-2,7.7) {Prague-Dejvice (S)};
\fill[recallmissing] (0,7.3100000000000005) rectangle (100,8.09);
\fill[recallten] (0.00000,7.3100000000000005) rectangle (55.10000,8.09);
\node[font=\scriptsize,text=white] at (27.55000,7.7) {55.1};
\fill[recalltwenty] (55.10000,7.3100000000000005) rectangle (82.82000,8.09);
\node[font=\scriptsize,text=white] at (68.96000,7.7) {82.8};
\fill[recallfifty] (82.82000,7.3100000000000005) rectangle (99.63000,8.09);
\node[font=\scriptsize,text=black!75] at (91.22500,7.7) {99.6};
\fill[recallhundred] (99.63000,7.3100000000000005) rectangle (99.99000,8.09);
\fill[recalltwohundred] (99.99000,7.3100000000000005) rectangle (100.00000,8.09);
\draw[gray!55,line width=0.3pt] (0,7.3100000000000005) rectangle (100,8.09);
\node[anchor=center,font=\footnotesize] at (115,7.7) {96.1\%};
\node[anchor=east,font=\footnotesize] at (-2,6.7) {Prague-Dejvice (I)};
\fill[recallmissing] (0,6.3100000000000005) rectangle (100,7.09);
\fill[recallten] (0.00000,6.3100000000000005) rectangle (54.93000,7.09);
\node[font=\scriptsize,text=white] at (27.46500,6.7) {54.9};
\fill[recalltwenty] (54.93000,6.3100000000000005) rectangle (82.46000,7.09);
\node[font=\scriptsize,text=white] at (68.69500,6.7) {82.5};
\fill[recallfifty] (82.46000,6.3100000000000005) rectangle (99.54000,7.09);
\node[font=\scriptsize,text=black!75] at (91.00000,6.7) {99.5};
\fill[recallhundred] (99.54000,6.3100000000000005) rectangle (99.99000,7.09);
\fill[recalltwohundred] (99.99000,6.3100000000000005) rectangle (100.00000,7.09);
\draw[gray!55,line width=0.3pt] (0,6.3100000000000005) rectangle (100,7.09);
\node[anchor=center,font=\footnotesize] at (115,6.7) {96.0\%};
\node[anchor=east,font=\footnotesize] at (-2,5.7) {London (S)};
\fill[recallmissing] (0,5.3100000000000005) rectangle (100,6.09);
\fill[recallten] (0.00000,5.3100000000000005) rectangle (18.41000,6.09);
\node[font=\scriptsize,text=white] at (9.20500,5.7) {18.4};
\fill[recalltwenty] (18.41000,5.3100000000000005) rectangle (33.59000,6.09);
\node[font=\scriptsize,text=white] at (26.00000,5.7) {33.6};
\fill[recallfifty] (33.59000,5.3100000000000005) rectangle (58.40000,6.09);
\node[font=\scriptsize,text=black!75] at (45.99500,5.7) {58.4};
\fill[recallhundred] (58.40000,5.3100000000000005) rectangle (80.39000,6.09);
\node[font=\scriptsize,text=black!75] at (69.39500,5.7) {80.4};
\fill[recalltwohundred] (80.39000,5.3100000000000005) rectangle (99.40000,6.09);
\node[font=\scriptsize,text=black!75] at (89.89500,5.7) {99.4};
\draw[gray!55,line width=0.3pt] (0,5.3100000000000005) rectangle (100,6.09);
\node[anchor=center,font=\footnotesize] at (115,5.7) {92.7\%};
\node[anchor=east,font=\footnotesize] at (-2,4.7) {London (I)};
\fill[recallmissing] (0,4.3100000000000005) rectangle (100,5.09);
\fill[recallten] (0.00000,4.3100000000000005) rectangle (19.32000,5.09);
\node[font=\scriptsize,text=white] at (9.66000,4.7) {19.3};
\fill[recalltwenty] (19.32000,4.3100000000000005) rectangle (34.86000,5.09);
\node[font=\scriptsize,text=white] at (27.09000,4.7) {34.9};
\fill[recallfifty] (34.86000,4.3100000000000005) rectangle (61.15000,5.09);
\node[font=\scriptsize,text=black!75] at (48.00500,4.7) {61.2};
\fill[recallhundred] (61.15000,4.3100000000000005) rectangle (82.41000,5.09);
\node[font=\scriptsize,text=black!75] at (71.78000,4.7) {82.4};
\fill[recalltwohundred] (82.41000,4.3100000000000005) rectangle (99.55000,5.09);
\node[font=\scriptsize,text=black!75] at (90.98000,4.7) {99.6};
\draw[gray!55,line width=0.3pt] (0,4.3100000000000005) rectangle (100,5.09);
\node[anchor=center,font=\footnotesize] at (115,4.7) {92.9\%};
\node[anchor=east,font=\footnotesize] at (-2,3.7) {New York (S)};
\fill[recallmissing] (0,3.31) rectangle (100,4.09);
\fill[recallten] (0.00000,3.31) rectangle (33.99000,4.09);
\node[font=\scriptsize,text=white] at (16.99500,3.7) {34.0};
\fill[recalltwenty] (33.99000,3.31) rectangle (55.67000,4.09);
\node[font=\scriptsize,text=white] at (44.83000,3.7) {55.7};
\fill[recallfifty] (55.67000,3.31) rectangle (81.32000,4.09);
\node[font=\scriptsize,text=black!75] at (68.49500,3.7) {81.3};
\fill[recallhundred] (81.32000,3.31) rectangle (95.91000,4.09);
\node[font=\scriptsize,text=black!75] at (88.61500,3.7) {95.9};
\fill[recalltwohundred] (95.91000,3.31) rectangle (99.99000,4.09);
\draw[gray!55,line width=0.3pt] (0,3.31) rectangle (100,4.09);
\node[anchor=center,font=\footnotesize] at (115,3.7) {93.5\%};
\node[anchor=east,font=\footnotesize] at (-2,2.7) {New York (I)};
\fill[recallmissing] (0,2.31) rectangle (100,3.0900000000000003);
\fill[recallten] (0.00000,2.31) rectangle (35.63000,3.0900000000000003);
\node[font=\scriptsize,text=white] at (17.81500,2.7) {35.6};
\fill[recalltwenty] (35.63000,2.31) rectangle (58.94000,3.0900000000000003);
\node[font=\scriptsize,text=white] at (47.28500,2.7) {58.9};
\fill[recallfifty] (58.94000,2.31) rectangle (84.62000,3.0900000000000003);
\node[font=\scriptsize,text=black!75] at (71.78000,2.7) {84.6};
\fill[recallhundred] (84.62000,2.31) rectangle (96.93000,3.0900000000000003);
\node[font=\scriptsize,text=black!75] at (90.77500,2.7) {96.9};
\fill[recalltwohundred] (96.93000,2.31) rectangle (99.95000,3.0900000000000003);
\draw[gray!55,line width=0.3pt] (0,2.31) rectangle (100,3.0900000000000003);
\node[anchor=center,font=\footnotesize] at (115,2.7) {94.3\%};
\node[anchor=east,font=\footnotesize] at (-2,1.7000000000000002) {Sydney (S)};
\fill[recallmissing] (0,1.31) rectangle (100,2.0900000000000003);
\fill[recallten] (0.00000,1.31) rectangle (19.78000,2.0900000000000003);
\node[font=\scriptsize,text=white] at (9.89000,1.7000000000000002) {19.8};
\fill[recalltwenty] (19.78000,1.31) rectangle (32.54000,2.0900000000000003);
\node[font=\scriptsize,text=white] at (26.16000,1.7000000000000002) {32.5};
\fill[recallfifty] (32.54000,1.31) rectangle (61.27000,2.0900000000000003);
\node[font=\scriptsize,text=black!75] at (46.90500,1.7000000000000002) {61.3};
\fill[recallhundred] (61.27000,1.31) rectangle (79.04000,2.0900000000000003);
\node[font=\scriptsize,text=black!75] at (70.15500,1.7000000000000002) {79.0};
\fill[recalltwohundred] (79.04000,1.31) rectangle (90.92000,2.0900000000000003);
\node[font=\scriptsize,text=black!75] at (84.98000,1.7000000000000002) {90.9};
\draw[gray!55,line width=0.3pt] (0,1.31) rectangle (100,2.0900000000000003);
\node[anchor=center,font=\footnotesize] at (115,1.7000000000000002) {92.0\%};
\node[anchor=east,font=\footnotesize] at (-2,0.7000000000000002) {Sydney (I)};
\fill[recallmissing] (0,0.31000000000000016) rectangle (100,1.0900000000000003);
\fill[recallten] (0.00000,0.31000000000000016) rectangle (21.07000,1.0900000000000003);
\node[font=\scriptsize,text=white] at (10.53500,0.7000000000000002) {21.1};
\fill[recalltwenty] (21.07000,0.31000000000000016) rectangle (35.44000,1.0900000000000003);
\node[font=\scriptsize,text=white] at (28.25500,0.7000000000000002) {35.4};
\fill[recallfifty] (35.44000,0.31000000000000016) rectangle (66.02000,1.0900000000000003);
\node[font=\scriptsize,text=black!75] at (50.73000,0.7000000000000002) {66.0};
\fill[recallhundred] (66.02000,0.31000000000000016) rectangle (83.86000,1.0900000000000003);
\node[font=\scriptsize,text=black!75] at (74.94000,0.7000000000000002) {83.9};
\fill[recalltwohundred] (83.86000,0.31000000000000016) rectangle (94.37000,1.0900000000000003);
\node[font=\scriptsize,text=black!75] at (89.11500,0.7000000000000002) {94.4};
\draw[gray!55,line width=0.3pt] (0,0.31000000000000016) rectangle (100,1.0900000000000003);
\node[anchor=center,font=\footnotesize] at (115,0.7000000000000002) {92.3\%};
\node[anchor=north,font=\footnotesize] at (50,-0.45) {Top-$K$ recall (\%)};
\end{scope}
\end{tikzpicture}
}
\caption{GNN validation ranking performance in the static (S) and increasing (I) regimes; bar labels show cumulative recall.}
\label{fig:gnn_accuracy}
\end{figure}

\paragraph{Discussion}
Figure~\ref{fig:gnn_accuracy} shows that all eight models learn the target ordering consistently, with pairwise accuracy ranging from $92.0\%$ to $96.1\%$ and comparable between the two traffic regimes within each city. The Prague-Dejvice models achieve the highest accuracy, consistent with this network being the smallest of all four. The lowest values occur for Sydney, which is the largest studied network, and London, where a lower accuracy may also reflect its specific topology. Overall, the accuracy remains above $92\%$ across all cities and traffic regimes, with a spread of only about four percentage points, indicating that the GNN learns reliably and preserves the relative segment-usage ordering irrespective of network size or traffic intensity.

Similarly to pairwise accuracy, top-$K$ recall also exposes several differences among the city networks. In Prague-Dejvice, the top $50$ positions already recover more than $99.5\%$ of the labelled segments. New York reaches $95.9$--$96.9\%$ at $K=100$ and almost $100\%$ at $K=200$; London reaches only $80.4$--$82.4\%$ at $K=100$ but approximately $99.5\%$ at $K=200$, while Sydney reaches $90.9$--$94.4\%$ at $K=200$. Thus, even the cities with comparable total road lengths can require different prefix sizes -- for example, London has a shorter total road length than New York but more segments and turn-restricted junctions, and requires a larger $K$ to achieve comparable recall. As the required prefix sizes notably differ across the networks, the results support calibrating $K_{\mathrm r}$ separately for each city and traffic regime.

\subsection{Optimization-level evaluation}
\label{subsec:results_optimization}

This experiment evaluates whether the proposed approach outperforms other methods on the considered CTR MILP under the operational $10$-second solver budget. For each city and traffic regime, we evaluate five $N_F$ values selected in Section~\ref{subsec:results_scalability}, using ten test scenarios ($1200$ episodes per method) for each $N_F$ value. The reduction parameter $K_{\mathrm r}$ is calibrated independently for each $(\text{city},\text{mode},N_F)$ combination. To determine it, we first weight each ranked segment by the fraction of vehicles routed through it in the labelled solution; we then choose $K_{\mathrm r}$ as the smallest number of top-ranked segments containing $90\%$ of the total weight in the GNN validation data. The detour factor $\lambda$ is the maximum of $2.0$ and the $99$th percentile of route travel times in the training solutions relative to the shortest travel time for the same OD pair. The SP routing provides a feasible initial solution for every instance. A separate run of FG with a $30$-second time limit, denoted FG$_{\mathrm{ref}}$, provides a reference solution for each episode.

We evaluate each method in terms of its relative objective difference from FG$_{\mathrm{ref}}$ and runtime. The difference is defined as $d_a=(z_a-z_{\mathrm{ref}})/|z_{\mathrm{ref}}|$, where $z_a$ is the best objective value found by the evaluated method and $z_{\mathrm{ref}}$ is the corresponding best objective value found by FG$_{\mathrm{ref}}$. We report the average difference across all episodes for each city and traffic regime. As for runtime, the total time includes computing the scores (where applicable) and constructing the reduced network, as well as constructing and solving the MILP. The one-time loading of a GNN model is excluded. We also separately report the Gurobi time spent for solving the MILP. For each of the two runtime metrics, we remove the lowest and highest $10\%$ of values and average the remainder.

\begin{table}[!t]
\centering
\caption{%
Optimization-level evaluation results.
}
\label{tab:pipeline_mip}
\small
\setlength{\tabcolsep}{5pt}
\renewcommand{\arraystretch}{1.05}
\begin{tabular*}{\linewidth}{@{\extracolsep{\fill}}l r r r r r r r@{}}
\toprule
\multicolumn{8}{@{}l}{Mean relative objective difference to FG$_{\mathrm{ref}}$ (\%)} \\
\addlinespace[2pt]
City & FG & GNN & LP & Random & SR & BR & SP \\
\midrule
\multicolumn{8}{@{}l}{Static regime} \\
\addlinespace[2pt]
Prague-Dejvice & 1.14 & \textbf{1.12} & 142.00 & 4.52 & 40.34 & 23.85 & 422.96 \\
London & 1.68 & \textbf{0.54} & 3.85 & 1.17 & 309.25 & 201.56 & 545.60 \\
New York & 106.53 & \textbf{0.67} & 16.48 & 4.50 & 67.10 & 5.12 & 545.35 \\
Sydney & 31.06 & \textbf{0.24} & 118.94 & 6.34 & 43.51 & 11.51 & 341.89 \\
\addlinespace[1pt]
\textbf{All cities} & 35.10 & \textbf{0.64} & 70.32 & 4.13 & 115.05 & 60.51 & 463.95 \\
\midrule
\multicolumn{8}{@{}l}{Increasing regime} \\
\addlinespace[2pt]
Prague-Dejvice & 2.85 & \textbf{2.06} & 250.26 & 6.53 & 39.86 & 22.19 & 658.67 \\
London & 3.09 & \textbf{1.22} & 1.25 & 1.27 & 245.14 & 174.18 & 517.71 \\
New York & 87.15 & \textbf{-1.24} & 45.30 & 4.23 & 56.56 & 3.91 & 508.18 \\
Sydney & 49.10 & \textbf{-0.61} & 88.51 & 8.78 & 38.15 & 9.35 & 299.07 \\
\addlinespace[1pt]
\textbf{All cities} & 35.55 & \textbf{0.36} & 96.33 & 5.20 & 94.93 & 52.41 & 495.91 \\
\bottomrule
\end{tabular*}
\vspace{6pt}
\footnotesize
\begin{tabular*}{\linewidth}{@{\extracolsep{\fill}}l r r r r r r r@{}}
\multicolumn{8}{@{}l}{Runtime over all test episodes} \\
\addlinespace[2pt]
Method & FG & GNN & LP & Random & SR & BR & SP \\
\midrule
Solve time (s) & 9.08 & 4.88 & 3.02 & 6.99 & 0.42 & 0.99 & $<0.01$ \\
Total time (s) & 9.60 & 8.43 & 4.24 & 7.63 & 0.52 & 1.13 & 0.05 \\
\bottomrule
\end{tabular*}
\renewcommand{\arraystretch}{1.0}
\end{table}

\paragraph{Discussion}
Table~\ref{tab:pipeline_mip} shows that GNN obtains the smallest average relative objective difference from FG$_{\mathrm{ref}}$ across all four city networks and both traffic regimes. The advantage ranges from a small margin on the small and moderate-sized networks (Prague-Dejvice static, London increasing) to larger values on the bigger and computationally harder networks (New York and Sydney, in both settings). A particularly strong result is the negative difference achieved by GNN in the increasing regime for New York and Sydney: the GNN-reduced model, under a $10$-second budget, finds better solutions than FG$_{\mathrm{ref}}$, which is given a three-times longer, $30$-second budget -- it indicates that the GNN-based reduction is highly informative and can yield substantial time savings. We further note that every GNN run returned at least one feasible solution improving on the SP initialization. We note in addition that the FG$_{\mathrm{ref}}$ is tightly bounded in most of the episodes: its primal-dual gap is no more than $1\%$ in $94.4\%$ of static and $84.2\%$ of increasing-regime episodes. Thus, the mean certified gaps of GNN, computed using the full-graph lower bounds from FG$_{\mathrm{ref}}$, are $0.83\%$ and $1.16\%$, respectively.
 
Several observations arise when comparing the proposed GNN-based ranking with its Random and LP ablations. First, the comparison with Random shows the proposed network-reduction procedure remains useful regardless of the predicting oracle: Random, indeed, performs substantially better than the fixed reduction methods in nearly all cases and even improves on the $10$-second FG runs for the large city networks. However, the GNN-based ranking plays a significant role: for \emph{all cities} it reduces the mean relative objective difference compared with Random, the best other method in both regimes, by $3.49$ and $4.85$ percentage points in the static and increasing regimes, respectively. Both differences are statistically significant (Holm-adjusted $p=0.00036$). Moreover, in cases of New York and Sydney (increasing regime) the GNN even surpasses the FG$_{\mathrm{ref}}$ baseline. We also note that, despite using the same ranking prefix budget, GNN retains only $45\%$ of the flow--segment pairs available in FG, compared with $61\%$ for Random, which is reflected in a lower solving time of $4.88$~s versus $6.99$~s; a part of Random's performance can therefore be attributed to its larger reduced network, which increases the chance of containing a good solution. Finally, the comparison of GNN with the LP ablation shows GNN's clear dominance; the values obtained by LP indicate that the fractional primal values of its relaxation provide little useful ranking information for the subsequent integer optimization (in contrast to \cite{Dilkina2009}, Section~\ref{subsec:related_learning}).

\subsection{Microscopic simulation evaluation}
\label{subsec:results_simulation}
 
The final experiment examines how improvements observed at the optimization level translate into traffic outcomes in a microscopic simulation. For every method, we recover vehicle routes from each episode's MILP solution obtained in Section~\ref{subsec:results_optimization} and replay the resulting assignments in SUMO~\citep{SUMO2018}.  For each $(\text{city},\text{mode},N_F)$ combination, all methods use the same scenarios, vehicle departure times, and SUMO settings.
 
Since the traffic state depends on factors beyond the initial routes, an important question is how to formulate a valid experiment demonstrating how an optimization result leads to simulated traffic performance. For example, without rerouting, vehicles cannot respond to conditions arising during simulation (e.g., accidents, queues, and delays), so initial routes alone cannot realistically determine the outcome. Conversely, frequent route updates repeatedly overwrite the initial routing, so the experiment rather measures the adaptation to current traffic rather than the influence of the initial routes. We therefore use the following procedure: we define the share $p$ of rerouted vehicles and run each method on every scenario for $p\in\{0,0.05,\ldots,0.50\}$. SUMO updates the routes of the selected vehicles every $30$ seconds, matching the episode length. The results are then averaged over $p$ as defined below. We consider a one-hour simulation horizon, matching the period covered by the $120$ routing episodes. Although only a share of dispatched vehicles complete their trips within this window, extending the horizon further would increasingly reflect the network's natural clearance after dispatch ends, as queues dissipate and delayed trips complete regardless of routing quality. This keeps the comparison focused on differences between methods rather than generic post-dispatch dynamics.
 
We evaluate traffic performance using the arrival rate and the average trip duration. The arrival rate $A(p)$ is the percentage of dispatched vehicles that reach their destination within the one-hour operational horizon, and the average trip duration $D(p)$ is computed over completed trips. Since duration is conditional on arrival, we instead report the ratio $R(p)=D(p)/A(p)$, in seconds per percentage point (s/pp) of arrival. We average both metrics over the eleven rerouting shares:
\[
\overline A=\frac{1}{11}\sum_{i=0}^{10}A(0.05i),
\qquad
\overline R=\frac{1}{11}\sum_{i=0}^{10}R(0.05i).
\]

We make two remarks on the interpretation of the simulation results. First, the evaluation scenarios are intended as traffic stress tests rather than representations of typical traffic conditions. We deliberately impose high traffic loads: each OD pair accounts for an average of about $500$--$900$ vehicles over the hour, while the average total number of vehicles per scenario (given the selected numbers of OD pairs) is about $21000$, $6300$, $7700$, and $6400$ for Prague-Dejvice, London, New York, and Sydney, respectively. Such loads, combined with the one-hour simulation horizon, are expected to lead to low arrival rates, and the resulting SUMO values should therefore not be interpreted as estimates of normal city traffic performance; instead, they provide a basis for comparing the routing methods under the same difficult conditions. Second, the individual results of the methods should be interpreted within the limitations of the experiment. The proposed MILP aims to minimize tiered segment costs, whereas the simulation metrics depend on vehicle interactions, departure times, rerouting, queues, and other traffic conditions that the optimization does not model directly, so an exact correspondence between the two should not be expected. Nevertheless, the experiment is intended to show that the optimization-level gains of the proposed method remain beneficial overall under the simulation settings.

\begin{table}[!t]
\centering
\caption{%
Microscopic simulation across rerouting shares. Cells report average arrival (\%) / duration-to-arrival (s/pp); metric bests are bold.
}
\label{tab:simulation}
\small
\setlength{\tabcolsep}{1.8pt}
\renewcommand{\arraystretch}{1.08}
\begin{tabular*}{\linewidth}{@{\extracolsep{\fill}}l c c c c @{\hspace{4pt}}|@{\hspace{1.8pt}} c@{}}
\toprule
Method & Prague-Dejvice & London & New York & Sydney & \textbf{All cities} \\
\midrule
\multicolumn{6}{@{}l}{\textbf{Static regime}} \\
\addlinespace[2pt]
FG & 1.11/\textbf{401.0} & \textbf{16.29}/47.1 & 2.37/589.6 & 17.45/43.8 & 5.98/\textbf{130.4} \\
GNN & 1.17/407.6 & 15.83/48.9 & 2.44/585.2 & \textbf{17.89}/\textbf{43.6} & \textbf{6.02}/131.8 \\
LP & \textbf{1.21}/518.2 & 15.94/48.6 & \textbf{2.81}/571.3 & 12.24/85.5 & 5.27/177.0 \\
Random & 0.89/514.8 & 16.20/\textbf{47.1} & 2.66/\textbf{509.8} & 17.18/46.2 & 5.87/136.1 \\
SR & 0.45/691.9 & 10.63/99.8 & 1.75/987.4 & 15.36/54.4 & 4.40/216.2 \\
BR & 0.46/678.5 & 13.46/68.6 & 2.56/593.7 & 16.74/48.2 & 5.17/172.2 \\
SP & 0.46/605.7 & 9.64/116.4 & 1.06/1586.5 & 11.43/62.8 & 3.54/255.5 \\
\midrule
\multicolumn{6}{@{}l}{\textbf{Increasing regime}} \\
\addlinespace[2pt]
FG & 0.84/482.5 & 13.81/54.7 & 2.59/533.6 & 16.53/43.7 & 5.81/130.4 \\
GNN & 0.97/454.7 & 13.80/54.9 & 2.64/513.1 & \textbf{18.04}/\textbf{40.1} & \textbf{6.12}/\textbf{123.2} \\
LP & \textbf{1.45}/\textbf{441.4} & 13.76/55.2 & \textbf{3.25}/493.6 & 13.09/70.6 & 5.67/155.8 \\
Random & 0.93/472.6 & \textbf{13.86}/\textbf{54.2} & 2.89/\textbf{458.1} & 16.13/48.7 & 5.85/133.7 \\
SR & 0.49/659.6 & 9.81/99.4 & 2.12/782.8 & 15.25/51.8 & 4.68/191.4 \\
BR & 0.49/635.7 & 11.78/75.3 & 2.86/492.9 & 16.85/44.9 & 5.39/155.7 \\
SP & 0.50/519.3 & 8.32/129.5 & 1.40/1165.9 & 11.47/57.8 & 3.71/230.0 \\
\bottomrule
\end{tabular*}
\renewcommand{\arraystretch}{1.0}
\end{table}

\paragraph{Discussion}
We first analyze how well the optimization-level results lead to the microscopic traffic performance in general. Table~\ref{tab:simulation} shows that the ordering from Section~\ref{subsec:results_optimization} is not maintained precisely across separate methods, but rather preserves a clear difference between groups of methods: FG and the ranking-based approaches demonstrate stronger simulation results, while the fixed reduction methods SR, BR, and SP tend to give lower arrival rates and larger duration-to-arrival ratios. In particular, GNN and LP obtain higher arrival rates in the \emph{All cities} results than every fixed reduction in both traffic regimes.

Among the individual methods, GNN demonstrates the strongest overall performance across cities: it shows the best arrival rate in both traffic regimes, while also ranking first in the increasing regime by $\overline R$ and second in the static regime by the same metric. We remark that the differences are statistically significant only for the increasing regime for both arrival rate (the best other method is Random) and duration-to-arrival ratio (the best other method is FG), with Holm-adjusted $p=0.00036$ for both comparisons. The GNN's performance is also reflected across individual cities: it stays among the top-three methods in most city/regime combinations -- five of eight by arrival rate and six of eight by $\overline R$. In addition, the GNN performs remarkably in Sydney, where it achieves the highest arrival rate ($17.89\%$ static, $18.04\%$ increasing) together with the lowest duration-to-arrival ratio in both the static ($43.6$ s/pp) and increasing ($40.1$ s/pp) regimes. Thus, the results confirm that GNN's stronger optimization-level performance also translates into stronger overall simulation performance across cities.

\section{Conclusion}
\label{sec:conclusion}

This paper addressed the computational difficulty of solving the CTR MILP over an urban road network within short operational time budgets. We proposed a network reduction procedure that exploits not only the topology of the road network but also information about the vehicles present in the current episode. A dedicated GNN ranks road segments separately for each OD flow, and the highest-ranked segments are retained to form a reduced network for solving the segment-flow MILP. The trained GNN thus narrows the search space available to optimization without predicting the routing itself, while the MILP ensures feasibility and determines the final routing under the same objective.

Unlike fixed reductions, which reuse the same reduced road network across all episodes, the proposed method constructs a separate reduced network for every episode: highly ranked segments are incorporated into complete origin-destination paths, and a local expansion adds neighboring segments. Since the construction is agnostic to the source of the segment scores, it can also use other rankings, e.g., those obtained from the linear programming relaxation of the considered MILP or even from a random ordering. This allows us to separate the contribution of the reduction procedure from that of the learned ranking.

We evaluated the proposed approach on four real urban road-network topologies under two traffic regimes. The scalability experiment identified ranges of OD flows pairs where the full-graph formulation is difficult to solve within the operational budget, but additional solver time yields substantially better solutions. Thus, these ranges provide both a motivation for network reduction and suitable training instances for the GNN. Across the eight trained models, the validation accuracy ranges from $92.0\%$ to $96.1\%$. Within each city and traffic regime, the same model remains effective across five consecutive problem sizes and previously unseen scenarios. At the optimization level, GNN achieves the smallest relative objective difference from the $30$-second full-graph reference in every city and traffic regime, averaging at $0.4$--$0.6\%$ across cities. The comparison with fixed reduction methods shows that the proposed reduction procedure improves solution quality already on its own; nevertheless, among the methods using this procedure, GNN demonstrates a noticeable advantage over both Random and LP, confirming the substantial contribution of the learned ranking. The final experiment, which evaluates how the optimization results translate into simulation performance, shows that GNN also obtains the highest average arrival rate in both traffic regimes.

An important limitation concerns the connection between the optimization model and traffic simulation. The considered MILP represents congestion through tiered segment costs, whereas SUMO additionally models traffic lights, queues, interactions between vehicles, lane changes, and rerouting over time. The optimization model therefore captures only a limited part of the resulting traffic dynamics. Although the simulation experiment shows that the gains observed at the optimization level remain useful under the considered simulation settings, it does not establish that every improvement in the MILP objective necessarily improves traffic performance. A natural direction for future work is to make the approach more responsive to the current traffic situation by incorporating information about traffic lights, queues, and observed travel times and by updating the ranking and routing as traffic conditions change. This would bring the optimization model closer to real-time traffic management while retaining the feasibility guarantees of the MILP.

\section*{Acknowledgements}

This research is supported by Czech Science Foundation (project no. 23-05575S and 25-18003S), by the European Union under the OP JAK project ROBOPROX (reg. no. CZ.02.01.01/\allowbreak{}00/\allowbreak{}22\_008/\allowbreak{}0004590), and by the Grant Agency of the Czech Technical University (project no. SGS26/\allowbreak{}188/\allowbreak{}OHK3/\allowbreak{}3T/\allowbreak{}37 and project no. SGS25/\allowbreak{}144/\allowbreak{}OHK3/\allowbreak{}3T/\allowbreak{}13).

\bibliographystyle{elsarticle-harv}
\bibliography{references}

\end{document}